# Learning Spectral Allocation: A Fractional Diffusion Framework for Adaptive Volumetric Segmentation

*Y.-H Shen[1] and T.-Q. Li[2,3,4,5]*

***Abstract*— We address adaptive computation in 3D medical image segmentation: instead of designing another backbone, we ask how much spectral mixing each network stage needs and let optimization answer. The answer takes a closed form. We derive FHEAT, a two-parameter operator family, from the discrete cosine transform (DCT) solution of a fractional heat equation. A fractional order α and a diffusion strength D govern the operator, and at D=0 it is exactly the identity. Reparametrized by the natural semigroup time $\tau = D^{\alpha}$, same-resolution instances of the family compose exactly, so any distribution of diffusion across same-resolution stages amounts to a single Sobolev-type regularizer of learned strength (D=0 adds none). This identity limit is the mechanism the framework rests on: the optimizer of each layer, not the designer, decides whether global mixing is needed and how sharp it should be. We instantiate FHEAT in a lightweight U-shaped architecture (Light-UNETR) paired with a Kolmogorov–Arnold mixer (KAN3D) with adaptive rational activations, yielding FHEAT-Seg. On three public benchmarks at 5% to 20% label rates, training produces gradient-driven spectral sparsification: seven of the eight stage-level operators drive D to zero, and the single survivor saturates at the sharpest admissible low-pass ( α ≈ 0.9) in the decoder layer that feeds the semi-supervised attention map. Since D=0 reduces FHEAT to the identity, the retired layers become exact shortcuts at inference, cutting FLOPs from 4.29G to 0.90G (a 79% drop) at 0.975M parameters. Freezing every stage except the active decoder matches or slightly exceeds the fully learnable model, so training appears to identify one useful site rather than distribute a mixing budget. Under a standard semi-supervised protocol, FHEAT-Seg reaches Dice scores of 90.47% (left atrium), 78.79% (Pancreas-CT), and 81.90% (BraTS 2019), ahead of five semi-supervised methods and the baseline Light-UNETR. The large variant also surpasses Light-UNETR-L under full supervision (Dice 93.09%, 85.11%, and 87.19%) with 2.851M parameters and 55.75G FLOPs. These results suggest that the allocation of spectral computation is a learnable property of optimization dynamics, not a manual design commitment.**

[1] School of Arts and Sciences, Fujian Medical University, Fuzhou, China (e-mail: shenyh@fjmu.edu.cn).
[2] School of Medical Imaging, Fujian Medical University, Fuzhou, China.
[3] Department of Medical Radiation Physics and Nuclear Medicine, Karolinska
University Hospital, Stockholm, Sweden.
[4] Department of Clinical Science, Intervention and Technology, Karolinska
Institute, Stockholm, Sweden.
[5] Department of Psychology, Stockholm University, Stockholm, Sweden.
* Corresponding author: T.-Q. Li (e-mail: tie-qiang.li@ki.se).

## I. INTRODUCTION

Volumetric medical image segmentation underpins quantitative analysis in clinical workflows such as radiotherapy planning and surgical navigation, yet dense voxel-level annotations remain prohibitively expensive to obtain. Two directions have consequently grown in parallel: lightweight architectures that keep the computation graph deployable, and semi-supervised learning (SSL) protocols that exploit unlabeled scans through pseudo-labeling and consistency regularization. Both have progressed substantially, but they are rarely decoupled in a controlled comparison, so it is usually unclear whether label efficiency should be credited to the backbone or to the protocol.

Existing lightweight networks prescribe a fixed global-mixing budget per stage: every stage carries the same dual-branch recipe, a local convolution path paired with a global attention path, and pays the full cost of global mixing whether or not its features require it. Under scarce labels this hand-crafted uniformity has a concrete cost: the model cannot shift its limited capacity toward the stages where the data actually call for global aggregation. We instead treat the allocation of global spectral computation as a learnable, per-layer optimization problem, and we give it a closed-form solution, derived from the fractional heat equation, in which the identity mapping is exactly reachable. Training can then retire unnecessary computation instead of carrying it by default. Learnable allocation of computation has appeared in other guises, including differentiable architecture search, sparsity-inducing gates, stochastic depth, and learned routing (Section II-C), but the question of where global mixing is truly needed has not previously been posed as an explicitly learnable, layer-level problem whose solution provably contains the identity.

We make this allocation learnable by equipping each stage with a single identity-containing operator family rather than a hand-crafted branch structure. The core of our design, the Fractional Heat Conduction Operator (FHCO), is a fractional-order spectral diffusion derived from the fractional heat equation: under Neumann boundary conditions its eigenmodes are exactly the type-II discrete cosine transform (DCT-II) basis, and the closed-form solution acts in the DCT spectrum as multiplication by $exp(-(D||k||)^{\alpha})$ . We embed this core operator inside a residual, gated module called FHEAT (Fig. 2(c)). The fractional order α controls how aggressively high frequencies are attenuated; the diffusion strength D scales the overall smoothing. Both are learnable per-stage scalars, so each stage interpolates continuously between the identity (no mixing) and strong low-pass aggregation. We stress that this identity limit is the design thesis, not a convenient by-product. Existing spectral operators (FNet, GFNet, AFNO) apply fixed or fully parameterized filters with no mechanism to identify and lock an exact identity, so they cannot be compiled to a shortcut at inference. Only an operator family that contains the identity turns the allocation of spectral computation into a learnable quantity and thereby enables learned sparsification.

Spectral diffusion alone is a linear diagonal transform, however, and must be paired with expressive feature mixing. We couple each FHEAT operator with KAN3D, a 3D Kolmogorov–Arnold mixer in which the fixed activation of a conventional feed-forward block is replaced by learnable rational functions. Since the spectral operator already supplies global, frequency-selective context, the mixer's job is channel-wise recombination rather than further spatial aggregation, and the learnable activations let each stage adapt its nonlinearity to its own feature statistics without widening the network.

Trained under scarce labels, the network answers the allocation question consistently. Across three public benchmarks and label rates from 5% to 20%, gradient-driven spectral sparsification emerges every time: the diffusion strength D goes to zero in all but one of the eight stage-level operators, collapsing them into identities, while the sole survivor saturates at the strongest admissible low-pass order ($\alpha \approx 0.9$). In all five settings tested, the survivor is the final decoder block, the same block whose operator output produces the attention map consumed by the SSL protocol. A controlled experiment that allows a learnable D only at this block matches or slightly exceeds the fully learnable model; the same control at a non-surviving block does not. Training thus identifies a single useful site rather than a distributed mixing budget. And because a zero D reduces FHEAT to the identity, every degenerate stage becomes a direct shortcut at inference, leaving a computation graph of 0.975M parameters and 0.90G FLOPs, 79% fewer FLOPs than the Light-UNETR backbone (Fig. 1).

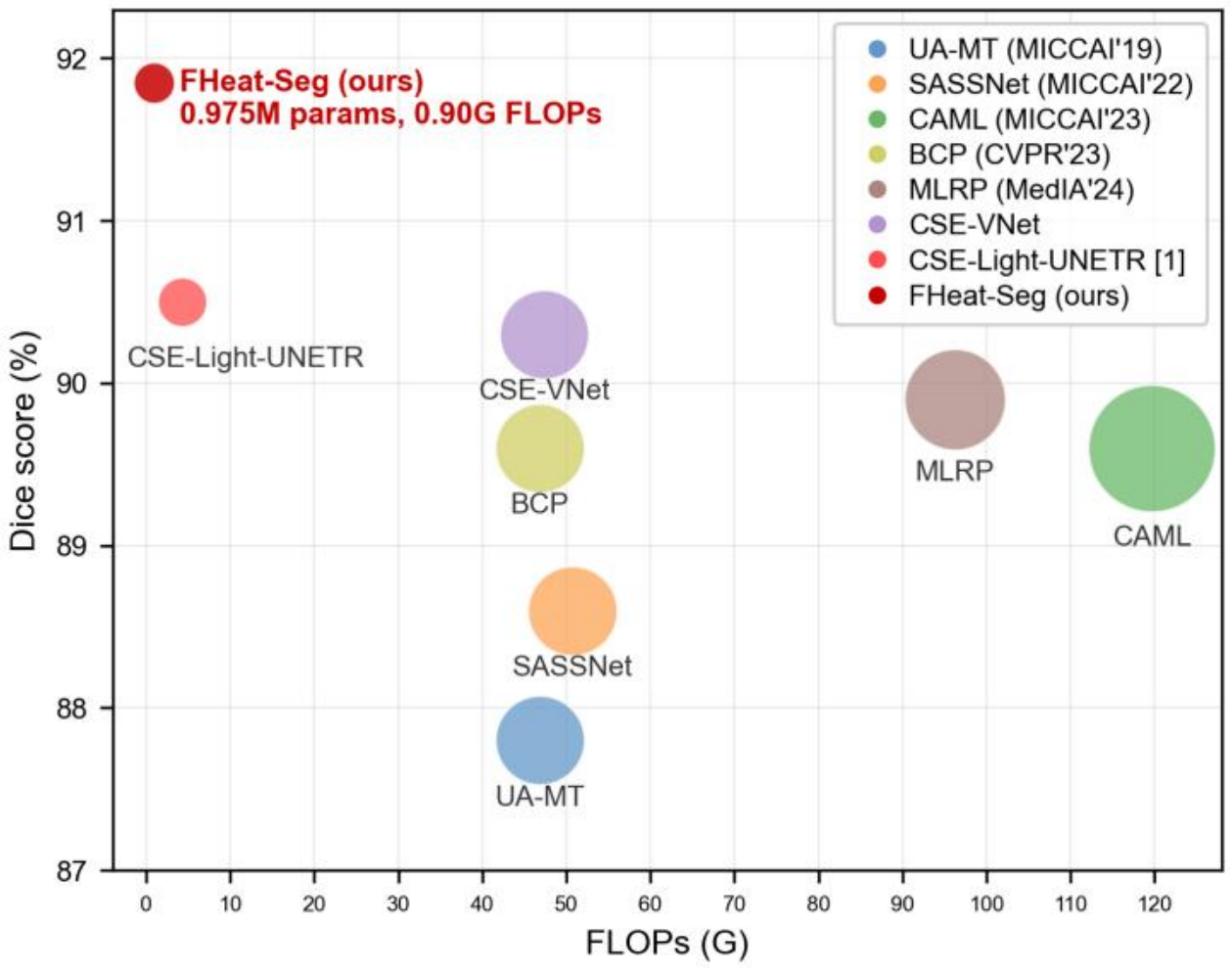

**Fig. 1.** Performance comparison between the proposed FHEAT-Seg and previous semi-supervised medical image segmentation methods on the LA dataset [1]. Circle size indicates the number of trainable parameters of each method.

We evaluate the framework on three public 3D benchmarks: left atrium (LA), Pancreas-CT, and BraTS 2019. Under the contextual synergic enhancement (CSE) semi-supervised protocol [2], which we keep unchanged throughout to isolate the backbone's contribution, FHEAT-Seg attains Dice of 90.47%, 78.79%, and 81.90% at 5–10% label rates, surpassing five semi-supervised methods, all trained on the same Light-UNETR backbone, as well as the backbone itself under an identical evaluation protocol. The large variant further outperforms Light-UNETR-L on fully supervised segmentation, with Dice of 93.09%, 85.11%, and 87.19% on the three benchmarks at lower parameter and FLOP counts (2.851M / 55.75G). Since these fully supervised results involve no semi-supervised component, they also indicate that the gains stem from the backbone itself.

We summarize the contributions of this work as follows:

We introduce the Fractional Heat Conduction Operator (FHCO), a two-parameter Neumann semigroup operator that is completely monotone, hence well-posed and non-amplifying (Remark 2), and that contains the identity exactly at zero diffusion strength (Remark 1), as a compilable per-stage gate for global spectral mixing. We instantiate it in a standard U-shaped encoder-decoder and pair it with a Kolmogorov–Arnold channel mixer (KAN3D) to obtain FHEAT-Seg, our empirical vehicle.

We prove that same-resolution, same-order gates form an exact semigroup: their diffusion times add, so any number of them equals a single gate at the summed time (Theorem 1). Analytically, and through a controlled non-medical simulation (Section III-F), we show that this redundancy alone does not explain sparsification, because weight decay applied to truly interchangeable gates favors spreading the diffusion budget rather than concentrating it (Remark 3). Empirically, training nonetheless selects a single representative: under scarce labels, all stage-level diffusion operators except one are driven to the identity, and in all five settings tested (three seeds each) the survivor coincides with the decoder block that produces the semi-supervised attention map. The result is an inference graph of 0.975M parameters and 0.90G FLOPs (79% fewer FLOPs than Light-UNETR) without accuracy loss.

We derive the first variation of the loss with respect to the natural semigroup time $\tau = D^\alpha$ and show that it is a fractional Dirichlet pairing between the incoming feature and the incoming gradient in the $\dot{H}^{\alpha/2}$ seminorm (Proposition 1): a stage is first-order useful if this pairing is strictly positive. Together with the ReLU-and-weight-decay absorption mechanism (Proposition 2), this identifies which stages survive instead of merely rationalizing that some do. The single-site ablations of Section IV-E agree with the ranking the pairing predicts (Block_Final useful; two other sites tested not), and the pairing itself is measured directly in a controlled synthetic experiment (Section III-F), where its sign at initialization correctly predicts the surviving site in all eight cases tested.

We evaluate on three benchmarks: state-of-the-art semi-supervised results under the shared CSE protocol at 5–20% label rates over five SSL methods and the Light-UNETR backbone, and higher fully supervised accuracy than Light-UNETR-L at lower cost. We also report a second, structurally different backbone (a 3D SegFormer with self-attention replaced by FHEAT gates, Section IV-B) that reproduces both the sparsification phenomenon and its exact-zero retirement, evidence that the operator-level claims are not restricted to Light-UNETR. Spectral analyses further reveal a division of labor in which the FHEAT blocks amplify high-frequency detail while the surviving attention-producing block performs low-pass aggregation.

## II. Related Work

### A. Lightweight 3D Medical Image Segmentations

Volumetric segmentation architectures have moved from heavy U-shaped CNNs toward compact designs that remain deployable under clinical compute budgets. Beyond the classic U-Net and V-Net families [3], [4], efficiency has been pursued along three main routes: depthwise-separable and group convolutions that cut per-block cost, self-configuring pipelines that optimize accuracy within a fixed compute envelope [5], and lightweight transformer or CNN–transformer hybrids that replace quadratic self-attention with low-rank or dual-branch attention [6], [7]. Light-UNETR [2] is representative of the last route: it interleaves local grouped convolutions with a low-frequency/high-frequency dual-branch attention module (LIDR) and reaches strong accuracy with a few million parameters. All of these designs, however, reduce the cost of each block while leaving the allocation of blocks hand-crafted: every stage receives the same global-mixing recipe regardless of need, so no capacity can be reclaimed from stages that do not require global context. Our work targets this next axis of efficiency: a parameterized operator family whose per-stage mixing strength is learned, so that training itself retains spectral computation only where it is needed.

### B. Semi-Supervised 3D Medical Image Segmentation

Semi-supervised segmentation seeks to exploit abundant unlabeled scans when voxel-level annotations are scarce. The dominant paradigm couples pseudo-labeling with consistency regularization: UA-MT [8] enforces uncertainty-aware consistency under input perturbations, BCP [9] perturbs predictions through bidirectional class-mixing, SASSNet [10] regularizes prediction maps with shape-aware constraints, CAML [11] enforces correlation-aware agreement between co-trained models, and MLRP [12] stabilizes mutual learning with reliability-filtered pseudo-labels. The CSE framework [2] combines Attention-Guided Replacement (AGR), which mixes labeled and unlabeled images in attention-selected regions, with Spatial Masking Consistency (SMC), and reports consistent gains when built on the Light-UNETR backbone. A persistent difficulty in this literature is that protocol and backbone are entangled, so improvements are hard to attribute to either. We therefore adopt the CSE protocol unchanged throughout this work and use it as a controlled environment in which the backbone is the only varied factor; all five competing SSL methods are compared under this shared-backbone setting.

### C. Frequency-Domain and Learnable-Activation Operators

Frequency-domain operators offer global context at a cost decoupled from spatial extent, and have been explored mainly in natural-image vision. FNet [13] replaces self-attention with fixed Fourier mixing, GFNet [14] learns arbitrary per-frequency filters, and AFNO [15] learns sparse token-mixing in the Fourier domain; KAN [16] and its rational-activation variants (KAT [17], later used in TK-Mamba [18]) make the activation functions themselves learnable. The learned filters in these operators are unconstrained: they carry no physical structure that guarantees well-posed smoothing and, more importantly, they give a stage no way to opt out. A fully parameterized spectral filter has no mechanism that identifies and locks an exact identity, so it cannot be compiled to a shortcut, even though gradient descent could in principle drive it arbitrarily close to one. AFNO's soft-thresholding of Fourier modes is the closest such mechanism in the prior literature, and it still leaves the retained modes as a fully parameterized, unlocked filter rather than an exact, checkable bypass, so training cannot answer whether a stage needs global mixing at all. FHEAT addresses both points: its frequency response is a two-parameter fractional heat kernel, completely monotone and hence positive definite in the squared frequency radius, with a built-in identity limit at zero diffusion strength, so the amount and the location of spectral mixing become learnable quantities.

Two recent operator-level baselines sharpen this distinction. SlimFormer-3D [19] also turns global mixing off by layer, but through post hoc gating factors selected from cross-layer activation similarity after training, not through a parameterization whose training-time optimum is an exact, checkable identity; its gates approximate redundancy, whereas ours certify it. AMBER-AFNO [20] replaces self-attention with AFNO-based frequency-domain mixing throughout a 3D segmenter, learning a fully parameterized spectral filter at every stage instead of asking, per stage, whether spectral mixing is needed at all. Neither baseline offers the two-parameter, completely monotone, exactly-D=0 family that lets a stage's own optimizer answer that question and lets the answer be compiled into a shortcut at inference. Our differentiator is this identity-containing, two-scalar, physically parameterized family, not simply the presence of a frequency mixer in a U-shaped backbone.

Two adjacent literatures contextualize this design. First, letting gradient descent decide whether a component contributes has a long history: differentiable architecture search relaxes discrete operator choices into learnable mixture weights [21], sparsity-inducing gates drive structural variables to exact zero [22], stochastic depth drops residual blocks at random during training [23], and mixture-of-experts models route tokens through learned gating functions [24]. Our question is related but distinct: rather than searching a discrete architecture space or pruning a fixed one, each stage declares, as a continuous training outcome, how much global mixing it warrants, and the identity-containing parameterization guarantees that “none” is an exactly reachable and directly deployable answer. Second, fractional-order diffusion operators have a long-standing lineage in classical PDE-based image processing, where fractional diffusion equations serve image denoising and enhancement [25]. FHEAT departs from that lineage in three ways: the fractional order is learned rather than hand-tuned, the operator is embedded per stage inside a trainable segmentation backbone, and its identity limit lets a stage retire itself entirely, which classical formulations do not provide.

Unlike hypernetworks [26], which condition weights on each input, the per-stage ($\alpha$, D) are learned once and held fixed,

which is exactly what makes the D = 0 bypass compilable ahead of inference.

## III. METHOD

### A. Overview

Section III-B–III-C develop the proposed framework independently of any specific backbone. Given a generic network layer $l$ with feature map $x_l$, we define a layer-wise spectral gate $G_l(D_l, \alpha_l)$ (the FHEAT operator) that governs how much global frequency-domain mixing that layer applies, paired with a KAN3D mixer for adaptive nonlinear channel recombination. Section III-F analyzes the optimization geometry that this gate induces. We then instantiate the framework empirically as FHEAT-Seg: a compact U-shaped network (overlapping patch embedding, four downsampling encoder stages with skip connections, and four decoder stages) that replaces every stage's hand-crafted mixing recipe with the learnable gate $G_l$, following the design principles of the lightweight backbone of [2] purely to obtain a fair, strong empirical baseline. The goal is a segmentation network that is both label-efficient and deployable under scarce annotations. For the semi-supervised experiments, the network is trained under the CSE protocol [2] (Section III-D), which we adopt unchanged and treat throughout as a fixed, controlled testbed rather than as a design constraint on the framework itself (Section II-B).

Two width configurations are used throughout. FHEAT-Seg stacks (1, 2, 3, 2) Light-UNETR blocks of 24, 48, 60, and 96 channels across its four encoder stages, with a mirrored decoder whose final block is the attention-producing Block_Final (Section III-C); each Light-UNETR block contains a depthwise 3×3×3 convolution and two FHEAT blocks (Fig. 2). FHEAT-Seg-L widens the design: a 32-channel stem and two plain convolutional stages (32 and 60 channels) precede the bottleneck, and five stages at 96, 150, 198, 150, and 96 channels (two encoder stages, the bottleneck, and two decoder stages, two Light-UNETR blocks each) carry the FHEAT blocks. Fig. 2 illustrates the architecture.

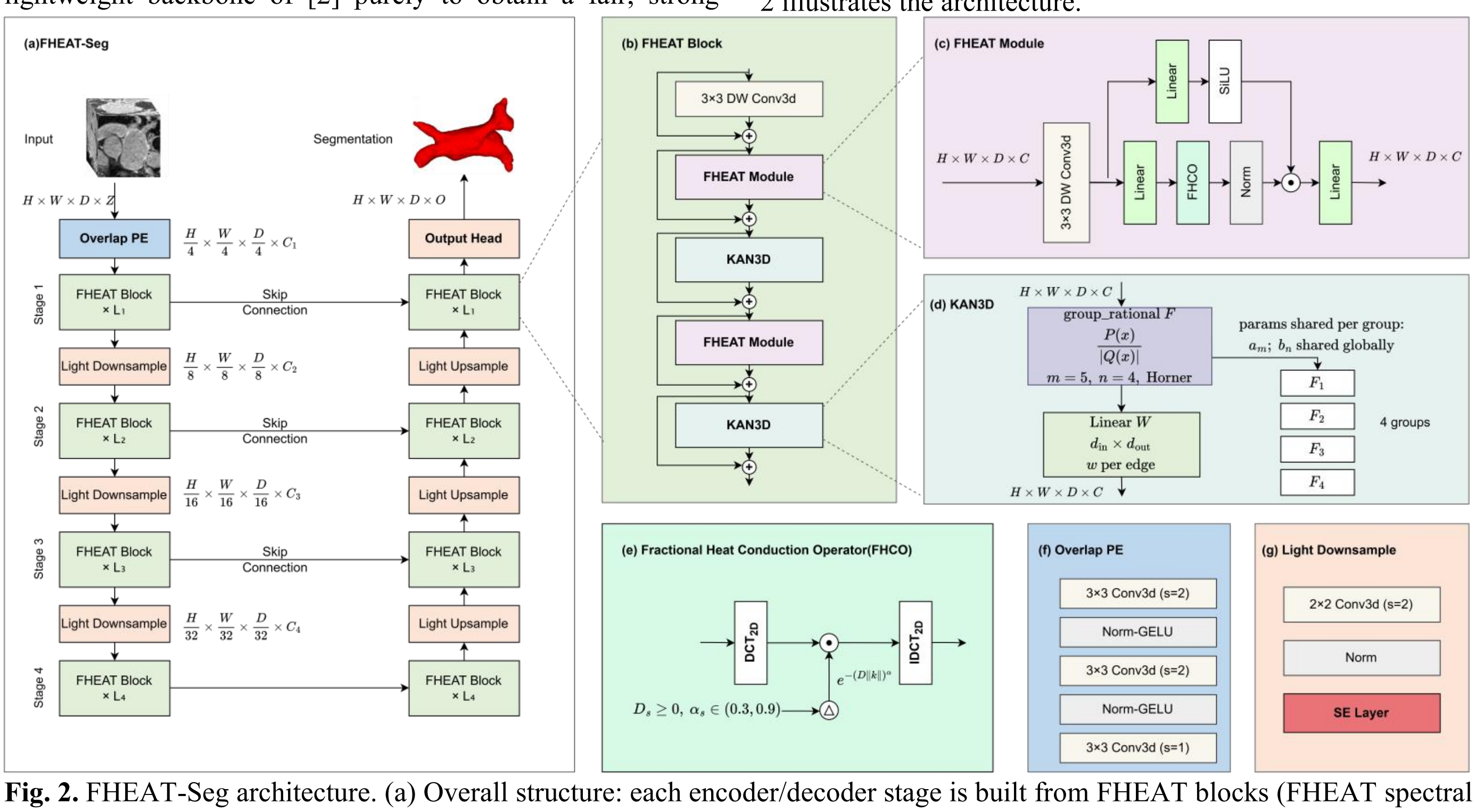

**Fig. 2.** FHEAT-Seg architecture. (a) Overall structure: each encoder/decoder stage is built from FHEAT blocks (FHEAT spectral diffusion + KAN3D mixing). (b) FHEAT block. (c) FHEAT. (d) KAN3D. (e) Fractional Heat Conduction Operator(FHCO). (f) Overlap patch embedding. (g) Light downsample.

### B. Fractional-Order Spectral Diffusion

Global context in a feature map amounts to aggregation across spatial scales, and spectral operators realize such aggregation by reweighting frequency coefficients. A fixed reweighting, however, commits every stage to one mixing behavior; we instead seek a parametric family in which both the shape and the amount of mixing are learnable per stage, and which contains the identity mapping, so that a stage can learn to forgo global mixing altogether.

We obtain such a family from the fractional heat equation. Consider a feature field $u$ on the stage domain $\Omega$ evolving as

$$\frac{\partial u}{\partial t} = -\kappa(-\Delta)^{\alpha/2}u, \quad \left.\frac{\partial u}{\partial n}\right|_{\partial\Omega} = 0 \tag{1}$$

where $\kappa$ is a diffusivity and  is the fractional order. Under the Neumann boundary condition, the eigenfunctions $(-\Delta)^{\alpha/2}$ of  are cosine modes, precisely the DCT-II basis, so the spectral basis is derived from the boundary physics rather than assumed. We adopt the spectral definition of the fractional Laplacian throughout, in which it is defined by eigenfunction expansion and therefore inherits exactly this cosine basis under Neumann boundary conditions [27]; this choice, rather than the nonlocal integral (Riesz) definition, is the natural one on a bounded voxel grid, where the DCT diagonalization yields the closed-form solution below. The normalized eigenmodes are

$$\varphi_k(x) = \prod_{a\in\{h,w,d\}} \cos\left(\frac{\pi k_a(2x_a+1)}{2N_a}\right), \quad k_a = 0, \ldots, N_a - 1 \tag{2}$$

Expanding $u(x,t) = \sum_k \hat{u}(k,t)\varphi_k(x)$ in this basis decouples (1) into independent scalar ODEs, $\frac{d\hat{u}(k,t)}{dt} = -\kappa \| k\|^{\alpha}\hat{u}(k,t)$,one per mode $k = (k_h, k_w, k_d)$ with per-axis DCT-II frequencies in $[0,\pi)$. Their closed-form solution is

$$\hat{u}(k,t) = \hat{u}_0(k)\exp\left(-\kappa t \| k\|^{\alpha}\right) \tag{3}$$

where $\| k \|$ is the Euclidean frequency radius. Reparametrizing the product $\kappa t$ as a single diffusion strength $D = (\kappa t)^{1/\alpha}$, the operator acts by pointwise multiplication of DCT-II coefficients with the closed-form kernel $\exp(-(D \| k \|)^{\alpha})$.

The spectral mapping

$$T_D: v \mapsto C^{-1}\left[e^{-(D\|k\|)^{\alpha}} \odot C(v)\right] \tag{4}$$

constitutes the Fractional Heat Conduction Operator (FHCO) (Fig. 2(e)).

Two per-stage scalars parameterize the response: the fractional order $\alpha$, which shapes how aggressively frequencies are attenuated, and the diffusion strength $D$, which scales the overall amount of smoothing. One parameter pair $(\theta_s, \delta_s)$ is shared by all blocks of stage $s$, so the eight stages learn 16 spectral scalars in total, bounded and initialized as

$$\begin{aligned} \alpha_s &= 0.3 + 0.6\,\sigma(\theta_s) \in (0.3,0.9), \\ D_s &= \mathrm{ReLU}(\delta_s) \geq 0, \\ \theta_s^{(0)} &= -2.0, \quad \delta_s^{(0)} = 0.03 \end{aligned} \tag{5}$$

It is convenient to name, once and for all, the quantity $\tau_s := D_s^{\alpha_s} \geq 0$ that this reparameterization makes natural: the kernel $\exp(-(D \| k \|)^{\alpha})$ inside (4) equals$\exp(-\tau \| k\|^{\alpha})$, so $\tau$ is the diffusion time of the underlying heat flow (3) and $D$ is only a coordinate on it. We use $D$ throughout the architecture description below because it is the quantity the implementation stores and initializes; Section III-F works directly in $\tau$, where the gates optimization geometry is best behaved.

Every stage therefore starts at $\alpha \approx 0.37$ with a mild, nonzero diffusion strength. The range $(0.3,0.9)$ is a design choice that keeps $\alpha$ inside the completely monotone regime while interpolating from a mild to a sharp low pass. The saturation of the learned $\alpha$ at this boundary, observed in Section IV, is interpreted in Section V. The bounded range is a practical choice to ensure well-posedness; we emphasize that the theoretical framework of Section III-F (Remark 1 and the Propositions built on it) does not require this bound, since those results concern the $D \to 0$ identity limit and the sign of $\delta$, not the range of $\alpha$. Section V explicitly diagnoses the saturation and proposes an unbounded ablation as future work, but the core sparsification phenomenon ($D \to 0$) is independent of this bound.

Given an input feature $x \in \mathbb{R}^{C\times H\times W\times Z}$, FHEAT first applies a depthwise $3 \times 3 \times 3$ convolution and a pointwise projection to $2C$ channels, split into a transformed branch $v$ and a gating branch. The FHCO then acts on v in the DCT-II domain. With $C$ and $C^{-1}$ denoting the separable 3D DCT-II and its inverse, implemented as matrix products with precomputed orthogonal cosine matrices along the three spatial axes, and with the kernel of (4) applied pointwise in the transform domain, the operator computes

$$\begin{aligned} y &= W_{\text{out}}(\mathrm{SiLU}(g) \odot \mathrm{GN}(C^{-1} \\ &\quad [\exp(-(D \| k \|)^{\alpha}) \odot C(v)])) \end{aligned} \tag{6}$$

Here $[v, g] = W_p(\mathrm{DWConv}_{3\times3\times3}(x))$ are the transform and gating branches produced by the pointwise projection, GN denotes group normalization, and $W_{\text{out}}$ projects the gated features back to $C$ channels; a residual connection is applied outside the block.

This parameterization provides three properties that unconstrained spectral filters do not. (i) Identity limit. As $D \to 0$, the kernel inside(4) tends to one for every frequency, so the entire transform reduces to the identity and can be replaced by a skip connection with unchanged outputs, a built-in "off switch" for the stage's global mixing. (ii) Well-posedness. For $\alpha \in (0,2]$, $s \mapsto \exp(-D^{\alpha}s^{\alpha/2})$ is completely monotone in the squared radius $s = \| k\|^2$ ; the kernel is therefore positive definite, and the operator performs well-posed smoothing rather than arbitrary per-frequency reweighting. (iii) Cost. With the separable matrix-product implementation, the transform costs $O(CHWZ(H + W + Z))$ per stage, linear in the channel count $C$; here $(H, W, Z)$ denote the spatial extents, with depth written $Z$ to avoid confusion with the diffusion strength $D$. An FFT-based fast DCT can reduce this to $O(CHWZ\log(HWZ))$ when needed.

### *C. FHEAT Block and KAN3D Mixing*

The FHEAT operator is a linear, frequency-diagonal transform: it reweights global context but cannot by itself recombine channels nonlinearly. Each stage therefore pairs FHEAT with a channel mixer, and rather than a fixed-activation feed-forward block, we use a mixer whose activation functions are themselves learnable.

Each FHEAT block consists of two residual sub-blocks,

$$x \leftarrow x + \mathrm{GN}(\mathrm{FHEAT}(x)) \tag{7}$$

$$x \leftarrow x + \mathrm{GN}(\mathrm{KAN3D}(x)) \tag{8}$$

where GN denotes single-group group normalization and the depthwise $3 \times 3 \times 3$ convolution is the first layer of the FHEAT operator (Section III-B). KAN3D is a 3D Kolmogorov--Arnold mixer: GN $\to$ $1 \times 1 \times 1$ convolution ($C \to rC$) $\to$ learnable rational activation $\to$ $1 \times 1 \times 1$ convolution ($rC \to C$), with expansion ratio $r = 1$. Following the Kolmogorov--Arnold principle of placing learnable univariate functions on the network's edges, the activation is a ratio of two polynomials with learnable coefficients (initialized to approximate GELU), so the mixer can adapt its nonlinearity to the feature statistics of each stage without widening the network. The final decoder block (Block_Final) additionally exposes an attention map

$$a = \frac{1}{C}\sum_c |y_c| \tag{9}$$

computed from the magnitude of its FHEAT output; this map is consumed by the semi-supervised protocol (Section III-D).

This pairing divides the labor cleanly: FHEAT supplies global, frequency-selective context, while KAN3D supplies adaptive pointwise recombination at negligible spatial cost (only $1 \times 1 \times 1$ convolutions). Section IV isolates the mixer's contribution with an equal-capacity GELU feed-forward control.

### D. Integration with the CSE Protocol

To evaluate the framework under label scarcity, we adopt the CSE semi-supervised protocol as a fixed, unmodified testbed rather than as a component of the proposed method: any semi-supervised protocol compatible with an attention read-out could in principle substitute for it, and we hold it constant precisely so that the backbone remains the only varied factor (Section II-B). CSE has two components: (a) Attention-Guided Replacement (AGR), in which an attention map identifies informative regions and the corresponding regions of an unlabeled image replace those of the labeled image, producing a mixed input supervised by a region-wise combination of the ground-truth and pseudo labels; and (b) Spatial Masking Consistency (SMC), in which a strongly masked view of each volume is enforced to predict consistently with its weakly augmented view.

In our network, the attention map consumed by AGR is produced by Block_Final (Section III-C) and is computed with a stop-gradient: gradients from the AGR mixing loss do not flow through the attention map into the backbone, although all segmentation losses continue to train the backbone through the segmentation outputs. Attention thus remains a read-only region-selection signal: the mixing objective can exploit the attention map but cannot reshape the segmentation features that generate it. Protocol and backbone stay decoupled by construction, a property we rely on in Section IV when attributing gains to the backbone.

### E. Inference Simplification and Efficiency Accounting

The diffusion strengths learned during training define the deployed computation graph. After training, each stage evaluates ; wherever it is zero, the FHCO path of that stage is replaced by the identity, while the surrounding depthwise convolution, projections, and gating are retained. Because $D_s$ is exactly zero whenever $\delta_s \leq 0$, the bypass is exact, not approximate, for every retired stage: the kernel inside(4) reduces to the frequency-independent constant 1, which the subsequent group normalization removes. We measured a maximum relative deviation of $1.5 \times 10^{-6}$(fp32) between the branch and an identity branch, i.e., pure floating-point rounding. In every semi-supervised model we trained, this rule removes the spectral transform from all but one stage (Section IV-B), reducing inference cost from 1.20G to 0.90G FLOPs for FHEAT-Seg and from 57.77G to 55.75G for FHEAT-Seg-L. We report inference-time computation graphs throughout: terms that do not participate in the graph are excluded from parameter counts, and all parameters/FLOPs for the proposed and reference models are measured under an identical protocol (fvcore, single-channel input). The per-stage scalars add sixteen parameters in total. For transparency, the reference implementation also carries per-stage frequency-embedding tensors inherited from [1][2] (0.23M for the small model) that participate in no computation. These tensors are excluded from all reported counts and are removed in our released code, so a reader re-implementing the method from this section obtains exactly the reported parameter counts.

### F. Optimization Geometry of the Identity Limit

The FHEAT gate contains the identity as an exact fixed point. This subsection records four facts about that family: it cannot amplify (Remark 2); same-resolution, same-order gates form a semigroup and therefore collapse to a single time (Theorem 1); that collapse does not by itself produce sparsity under weight decay (Remark 3); the first variation of the loss in the semigroup time is a fractional Dirichlet pairing whose sign selects the site (Proposition 1), after which ReLU and weight decay compile an unfavourable sign into an exact zero (Proposition 2). Throughout, $\mathcal{L}$ is the training loss and $C$ is the orthonormal DCT-II of Section III-B. The discrete Neumann fractional Laplacian $(-\Delta)^{\alpha/2}$ is the diagonal operator with eigenvalue $\| k\|^{\alpha}$ on mode $k$. None of these statements names the surviving layer; that is an empirical outcome, tested by the single-site ablations of Section IV-E, by the training-dynamics instrumentation of Section IV-B, and by the controlled simulation in Fig. 3.The simulation uses 1D signals on a length-256 grid in the DCT-II basis with a fixed, shared $\alpha = 0.75$, optimizing $D = \mathrm{ReLU}(\delta)$ by AdamW (lr 0.02–0.03, decoupled weight decay $10^{-3}$ on $\delta$) for 6,000 iterations.

**Remark 1 (Exact identity).** Let $T_D$ denote the FHCO defined in (4),i.e., the spectral map inside (6).Then $T_0 = \mathrm{Id}$ for every input and every $\alpha \in (0,2]$: the multiplier is identically one, so $T_0 = C^{-1}C$ by orthogonality of (2). At $D = 0$ the FHCO reduces to the identity, and the FHEAT block collapses to the affine bypass around the transform,

$$y = W_{\text{out}} \quad (\mathrm{SiLU}(g) \odot \mathrm{GN}(v)) \tag{10}$$

and the inference replacement of Section III-E is exact (maximum relative deviation $1.5 \times 10^{-6}$ in fp32).

**Remark 2 (Non-amplification).** The eigenvalues $m_D(k) = \exp(-(D \| k \|)^{\alpha})$ lie in $(0,1]$, with $m_D(0) = 1$. Hence $\| T_D\|_{2\to 2} = 1$. On the fluctuation subspace $k \neq 0$,

$$\| T_D \| = \exp \quad (-(D \| k\|_{\min})^{\alpha}) < 1 \quad (D > 0) \tag{11}$$

a strict contraction that deepens in $D$. Equivalently, $s \mapsto \exp(-D^{\alpha} s^{\alpha/2})$ is completely monotone on the squared radius $s = \| k\|^2$ for $\alpha \in (0,2]$, so the kernel is positive definite. The gate is a well-posed smoother: it cannot manufacture energy at any scale. Unconstrained filters (GFNet, AFNO) have no such bound.

**Theorem 1 (Semigroup collapse).** Fix a DCT grid and a shared order $\alpha$. Write $\tau := D^{\alpha} \geq 0$ and let $T_{\tau}^{\alpha}$ denote the FHCO with multiplier $m_{\tau}(k) = \exp(-\tau \| k\|^{\alpha})$. Then

$$T_{\tau_1}^{\alpha} \circ T_{\tau_2}^{\alpha} = T_{\tau_1+\tau_2}^{\alpha}, \quad T_0^{\alpha} = \mathrm{Id}. \tag{12}$$

*Proof.* Orthogonality of $C$ gives $T_{\tau_1}^{\alpha} \circ T_{\tau_2}^{\alpha} = C^{-1}\mathrm{diag}(m_{\tau_1} m_{\tau_2})C$. The multipliers multiply pointwise,

$$m_{\tau_1}(k)\, m_{\tau_2}(k) = \exp \quad (-(\tau_1 + \tau_2) \| k\|^{\alpha}) = m_{\tau_1+\tau_2}(k) \tag{13}$$

Thus any finite composition of same-grid, same-$\alpha$ gates is a single gate at $\tau_{\text{tot}} = \sum_i \tau_i$, or $D_{\text{eff}} = (\sum_i D_i^{\alpha})^{1/\alpha}$. Within one FHEAT-Seg stage this is an identity, not an approximation: the blocks of stage $s$ share $(\theta_s, \delta_s)$. Across stages both hypotheses fail --- the grids differ, and each stage learns its own $\alpha_s$ --- so the eight stage-level gates are not interchangeable, and times must not be added across the U-Net. Theorem 1 only certifies the redundancy already built into each stage.

**Remark 3 (Collapse does not imply sparsity).** Theorem 1 says that interchangeable gates may be replaced by one. It does not say that training will pick a sparse representative. Under the sole constraint $\sum_s D_s^\alpha = \tau_{\text{tot}}$ and a quadratic penalty $\sum_s D_s^2$, stationarity requires $D_s^{2-\alpha}$ equal on the active set, hence the symmetric split

$$D_s = (\tau_{\text{tot}}/S)^{1/\alpha} \quad (14)$$

For $\alpha < 1$ this interior point is cheaper than any sparse corner: with $\alpha = 0.75$, $S = 8$, $\tau_{\text{tot}} = 2$, one has $\sum_s D_s^2 \approx 0.20$ versus 6.35 for concentration at a single site. Fig. 3(a) reproduces the dynamics: eight identical same-resolution gates fitted to one target remain active and near-equal ( $D = 0.248 \pm 0.011$ ). Weight decay on redundant gates therefore spreads the budget. The sparsity observed empirically cannot be blamed on Theorem 1 plus weight decay; it requires the stages to differ, which is what Proposition 1 tests.

**Proposition 1 (First variation).** Let $v$ be the tensor entering the spectral multiply in (6), let $z = T_\tau v$ be that multiplication's output (before group normalization and gating), and let $g := \partial\mathcal{L}/\partial z$. With $\hat{v} = Cv$ and $\hat{g} = Cg$ real,

$$\frac{\partial\mathcal{L}}{\partial\tau} = -\sum_{k\neq 0} \| k\|^\alpha\, m_\tau(k)\, \hat{v}(k)\, \hat{g}(k), \quad (15)$$

which is finite and continuous at $\tau = 0$. In particular

$$\left.\frac{\partial\mathcal{L}}{\partial\tau}\right|_{\tau=0} = -\langle v, (-\Delta)^{\alpha/2} g\rangle = -\langle (-\Delta)^{\alpha/4} v, (-\Delta)^{\alpha/4} g\rangle \quad (16)$$

Write $P := \langle (-\Delta)^{\alpha/4} v, (-\Delta)^{\alpha/4} g\rangle$ for the $H^{\alpha/2}$ pairing. Increasing $\tau$ from 0 strictly decreases $\mathcal{L}$ if and only if $P > 0$, i.e., if and only if the incoming feature and the incoming gradient are positively aligned in the $H^{\alpha/2}$ seminorm. Complementary slackness along this coordinate is therefore

$$P \leq 0 \iff \tau = 0 \text{ is a one-sided local minimizer} \quad (17)$$

$$P > 0 \iff \text{the gate should leave } \tau = 0 \quad (18)$$

To first order, the number of surviving stages is the number of stages with $P > 0$. The single-site ablations of Section IV-E agree with the ranking the pairing would predict (Block_Final useful). Fig. 3(b) already confirms the mechanism on a synthetic ground truth: the sign of $P$ at initialization predicted the unique survivor in all eight planted sites.

Proposition 2 (Absorption). The implementation optimizes $D = ReLU(\delta)$ rather than $\tau$, with

$$\delta \leftarrow \delta - \eta\left(\frac{\partial\mathcal{L}}{\partial\delta} + \lambda\delta\right) \quad (19)$$

The chain rule is singular at the origin for $\alpha \in (0,1)$ (the whole range used here), which is why Proposition 1 is stated in $\tau$. Once $\delta < 0$, $D = 0$ and the update is the contraction $\delta \leftarrow (1 - \eta\lambda)\delta$: $\delta = 0$ is absorbing. Combined with Proposition 1, a negative pairing $P < 0$ at $\tau = 0$ drives $\delta$ below zero and keeps it there; the gradient path closes at the boundary, and decoupled weight decay only contracts  toward zero from below, so no subsequent training signal can push $\delta$ back across. Retirement is therefore permanent for the rest of training, short of an external perturbation (cf. Fig. 5(g)).This is standard active-set identification for a hidden nonnegativity constraint [28]. It is specific to $D = \text{ReLU}(\delta)$ under decoupled weight decay --- a softplus or exponential coordinate would not lock --- and it is a compiler, not a cause. The empirical fact is that seven of eight pairings are unfavourable under the present protocol.

**Remark 4 (Sobolev reading).** The explicit multiplier $m_\tau(k) = e^{-\tau\|k\|^\alpha}$ agrees to first order in $\tau \| k\|^\alpha$ with the resolvent $1/(1 + \tau \| k\|^\alpha)$ of the Tikhonov problem

$$\min_u \frac{1}{2} \| u - v\|^2 + \tau\, |u|^2_{H^{\alpha/2}} \quad (20)$$

since $e^{-x}$ and $1/(1 + x)$ share the expansion $1 - x + O(x^2)$. The operators coincide only as $\tau \to 0$; $T_\tau$ is the exact heat flow, which is why Theorem 1 is an identity. The useful reading remains: raising $\tau$ at a stage is, to leading order, strengthening an $H^{\alpha/2}$ penalty on that stage's features, and $D = 0$ means adding no such penalty. FHEAT is a per-stage Sobolev regularizer whose strength is learned, not an unconstrained Fourier filter.

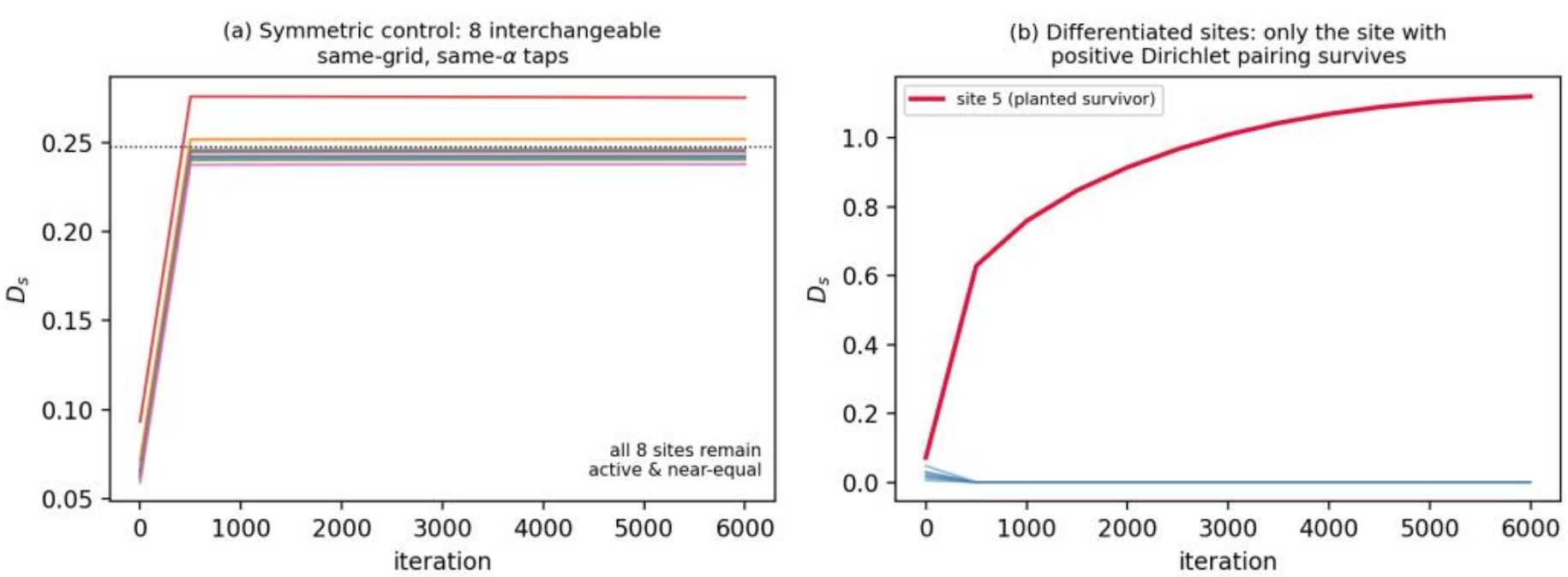

**Fig. 3.** Controlled, non-medical test of Theorem 1 and Proposition 1; no medical data or training involved. (a) Symmetric control: eight redundant, same-resolution gates jointly fit one shared low-pass target. Consistent with Remark 3, all eight remain active with $D = 0.248 \pm 0.011$ (mean $\pm$ SD): weight decay does not concentrate the budget onto genuinely interchangeable gates. (b) Differentiated sites: seven gates have a target equal to their input (pairing exactly 0 at $D = 0$); one (site 5, red) has a genuine low-pass target (pairing $\approx 1609$). Only the planted site survives (final $D = 1.12$ vs. exactly 0 at the other seven), and the sign of the pairing at initialization predicts survival at all eight sites, exactly as Proposition 1 predicts.

## IV. Experiments

### *A. Experimental Setup*

Datasets. We evaluate three public 3D benchmarks. (i) LA [1]: LGE-MRI scans; following common practice [9], we use 80 volumes for training and the remaining 20 for testing; 4 and 8 labeled scans correspond to 5% and 10% label rates. (ii) Pancreas-CT [29]: following the FUSSNet preprocessing and split [30] used by [2], 62 volumes for training and the remaining 20 for testing, with 6 and 12 labels (≈10% and ≈20%). (iii) BraTS 2019 [31]: 250, 25, and 60 volumes for training, validation, and testing, with 25 labels (≈10%).

Metrics. Dice score (Dice), Jaccard index (Jac), 95% Hausdorff distance (95HD), and average symmetric surface distance (ASD); 95HD and ASD are reported in voxels.

We conduct all experiments on NVIDIA 5090 GPU. Semi-supervised models are trained under the CSE protocol for 16,000 iterations (batch of two labeled + two unlabeled volumes; AdamW, weight decay, base learning rate 0.01 with cosine decay); the fractional-order parameters use a learning-rate multiplier, the diffusion strengths the base rate, and the KAN rational coefficients a multiplier. Following the reference protocol, model selection is performed on validation data held out from the training set, and all reported scores are computed on the separate test sets.

Unless stated otherwise, the improvement of FHEAT-Seg over the strongest baseline in each comparison is significant under a paired Wilcoxon signed-rank test across test cases.

Baselines. We compare against five semi-supervised methods (UA-MT [8], SASSNet [10], CAML [11], BCP [9], and MLRP [12]) and against the Light-UNETR backbone. All methods are evaluated under the same Light-UNETR backbone and the official splits, so the backbone is the only varied factor in the comparison. The fully supervised comparisons follow the same protocol [2]. For qualitative comparison we additionally re-implemented all five SSL methods on this backbone under an identical protocol. We report the shared-backbone numbers precisely because they are the only numbers that isolate the contribution of the backbone from that of the SSL method [8]–[12].

### *B. Learned Spectral Allocation*

Table I reports the learned diffusion strengths read directly from the checkpoints of every semi-supervised model trained under the final protocol. In all five headline models, spanning three datasets and label rates from 5% to 20%, and in all five protocol-matched variants, exactly one of the eight stage-level operators retains $D > 0$; the other seven are learned negative and collapse to identities. The survivor is the final decoder block in every setting, the block whose FHEAT output produces the attention map consumed by AGR. Its fractional order saturates at the strongest admissible low-pass ($\alpha \approx 0.90$), whereas the retired stages sit at $\alpha \approx 0.60$–$0.75$. The assignment is therefore consistent in location and in shape. Every setting was repeated with three independent random seeds; the surviving block was Block_Final in all 15 runs. Table I reports a representative seed per setting.Under fully supervised training the same sparsification tendency holds (typically zero or one surviving stage), but the survivor's location is no longer concentrated at the attention-producing block (e.g., a mid-decoder stage on LA, none on Pancreas-CT). Spectral sparsification is thus a general behavior of the operator family; its localization at the attention interface is specific to semi-supervised training.

TABLE I. LEARNED SPECTRAL ALLOCATION OF THE SEMI-SUPERVISED MODELS (CHECKPOINT VALUES). "SURVIVOR" = THE UNIQUE STAGE WITH $D > 0$; ALL SEVEN OTHER STAGES LEARNED $D = 0$. CORRESPONDS TO THE SUMMARY STATISTIC REPORTED IN SECTION IV-B.

| **Model (setting)** | **Survivor** | **D** | **α** |
|---|---|---|---|
| LA, 5% labels | Decoder Block_Final | 1.224 | 0.900 |
| LA, 10% labels | Decoder Block_Final | 0.968 | 0.899 |
| Pancreas-CT, 10% | Decoder Block_Final | 0.496 | 0.897 |
| Pancreas-CT, 20% | Decoder Block_Final | 0.693 | 0.900 |
| BraTS 2019, ≈10% | Decoder Block_Final | 0.418 | 0.900 |

The survivor's D also varies considerably across settings (Table I): 1.224 and 0.968 on LA at 5% and 10% labels, 0.496 and 0.693 on Pancreas-CT at ≈10% and ≈20%, and 0.418 on BraTS 2019, the most heterogeneous of the three tasks in both anatomy and label count. BraTS carries the lowest surviving D, consistent with a hypothesis that a harder, more heterogeneous segmentation target benefits from a gentler low-pass at the attention-producing block, since an aggressive one risks smoothing away the boundary detail multi-region tumor segmentation depends on; under this reading, D would trade off against target heterogeneity rather than against dataset size, since Pancreas-CT and BraTS have comparable case counts but different D.

The training dynamics on LA at 5% labels (plotted for the equal-capacity FFN variant, whose per-iteration checkpoints were retained) follow a consistent pattern: all diffusion strengths start from the shared 0.03 initialization, the seven retired operators reach exactly zero by the first recorded checkpoint, and the surviving operator's D grows while its α approaches 0.9, the same endpoint the KAN3D model reaches (Table I). The learned per-stage spectral response $\exp(-(D\|k\|)^\alpha)$ of the headline model follows the same division: a strong low-pass at the surviving stage and the identity everywhere else, as detailed later in this section.

Cross-Architecture Generalization. To test whether gradient-driven spectral sparsification is a property of the operator rather than of the Light-UNETR backbone, we ported the FHEAT gate (FractionalHeat3D, same code) onto a 3D SegFormer, a hierarchical Transformer architecture that differs substantially from Light-UNETR's CNN–Transformer hybrid. Self-attention in every encoder TransformerBlock is replaced by the FHEAT spectral operator, and the all-MLP decoder receives one additive FHEAT gate per stage, with the final decoder gate producing the CSE attention map. The ported model (4.95M parameters) is trained under the identical CSE protocol on LA at 5% labels for 16,000 iterations. The spectral sparsification phenomenon reproduces on this architecturally distinct backbone: seven of the eight stage-level diffusion operators converge to exactly $D = 0$, and the sole survivor is the deepest decoder stage (dec4), with D growing monotonically from 0.050 to 0.120 over training. The survivor's position differs from FHEAT-Seg (Block_Final), which is expected, since the gradient signal that drives D growth is architecture-dependent; the qualitative pattern (unique survivor plus exact-zero retirement) is architecture-independent. On the test set, the FHEAT-ported SegFormer3D achieves Dice = 0.9002 (Jaccard 0.8192, 95HD 5.33, ASD 1.79), compared to 0.8817 (0.7899, 7.13, 2.20) for the plain SegFormer3D control trained under the same protocol: a +1.85 Dice-point gain from the FHEAT framework on an independent backbone.

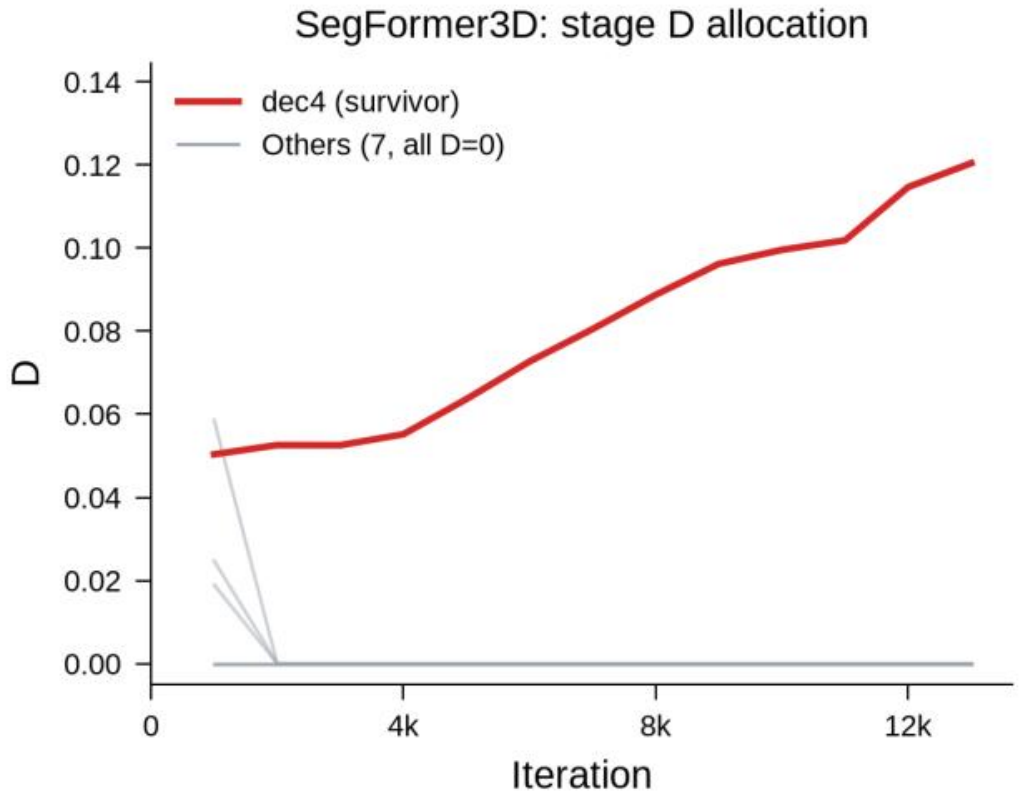

**Fig. 4.** Cross-architecture validation on SegFormer3D: stage D allocation. dec4 is the sole survivor (D 0.050→0.120); all 7 others converge to exactly D = 0. Test Dice: 0.9002 (+FHEAT) vs. 0.8817 (plain control).

The exactness and permanence of the zeros admit a precise dynamical reading. With $\mathcal{P}$ the Euclidean projection onto the feasible set $\mathcal{F}$ ,each raw diffusion parameter follows the subgradient dynamics up to the optimizer's weight-decay term. Two consequences follow. (i) Finite-time absorption. A stage whose loss favors removing spectral mixing ($P < 0$ as $\delta \to 0^{-}$) crosses $\delta = 0$ in finitely many steps and, because the indicator kills the gradient path at the boundary, cannot re-enter. This mirrors the finite-time active-set identification of projected-gradient methods [28], except that here the absorption is irreversible rather than re-activatable. (ii) Stability of the lock. The decoupled weight decay of AdamW only contracts a negative toward zero without changing its sign, so the identity state persists and the inference bypass of Section III-E remains stable for the rest of training. No explicit sparsity penalty appears anywhere in this mechanism: the loss landscape itself decides which stages cross the boundary, and the projection dynamics make each crossing exact and permanent. This is why the seven retired stages already sit at exact zero at the first recorded checkpoint and never recover (Fig. 5(a)).

Fig. 5(d) analyzes the bottleneck block residual (block output minus input) in the radial frequency domain: 33.3% of the FHEAT blocks residual energy lies at high normalized radii ($r > 0.6$), versus 0.2% for the LIDR block of Light-UNETR, whose dual-branch attention operates on downsampled features and injects almost purely low-frequency content. Read together with the learned low-pass response of the surviving operator (Fig. 5(c)), this suggests a division of labor: the FHEAT blocks preserve and inject high-frequency detail, while the single surviving spectral operator performs low-pass aggregation. The same low-pass operation also yields more focused attention: the attention maps of FHEAT-Seg concentrate 44.8% of their energy within the target region on LA, versus 3.4% for Light-UNETR (18.4% vs. 11.6% on Pancreas-CT; 22.5% vs. 1.3% on BraTS 2019; Fig. 6). Formally, with $a_v = (1/C)\sum_c |y_{c,v}|$ the stop-gradient attention magnitude at voxel $v$ and $\Omega$ the ground-truth foreground, the concentration ratio is $\sum_{v\in\Omega} a_v / \sum_v a_v$, computed per test case and averaged (LA: all 20 test cases, standard deviation 6.1 vs. 1.2 percentage points for FHEAT-Seg and Light-UNETR, respectively; Pancreas-CT: all 20 test cases; BraTS: all 60 test cases). Because the attention map is stop-gradient (Section III-D), we read this as a consistent by-product of the learned low-pass aggregation rather than as its cause. The Block_Final-only control of Section IV-E confirms that placing the spectral budget at exactly this block is sufficient.

C. Semi-Supervised Comparison

The allocation learned in Section IV-B defines the inference graph whose accuracy we now measure: one surviving low-pass operator at the attention-producing block, and seven retired stages running as free shortcuts. The gains reported below are the downstream consequence of that learned assignment. Table II reports the lower label setting (LA-4 at 5%, Pancreas-CT-6 at ≈10%). FHEAT-Seg achieves 90.47% Dice on LA and 78.79% on Pancreas-CT, surpassing the CSE-trained Light-UNETR backbone (89.53% / 73.77%) and all five SSL methods; the boundary metrics improve accordingly on LA (95HD 4.60 vs. 6.14, ASD 1.78 vs. 1.88). Table III reports the higher setting (LA-8 at 10%, Pancreas-CT-12 at ≈20%, BraTS-25 at ≈10%). The ordering is preserved at the higher label rates: FHEAT-Seg ranks first on Dice and Jaccard on every dataset of both tables, and the advantage extends to boundary metrics, where it attains the best 95HD and ASD in every setting of both tables, with one exception — 95HD on Pancreas-CT-12, where CSE-Light-UNETR remains ahead (8.35 vs. 8.69).Across three random seeds, the standard deviations of FHEAT-Seg are smaller than the corresponding margins over the CSE-trained Light-UNETR backbone in every setting of both tables, and the Dice and Jaccard improvements are significant (paired Wilcoxon signed-rank test across test cases, $p < 0.05$).

TABLE II. Semi-supervised segmentation at the lower label rate (LA: 4 labels, 5%; Pancreas-CT: 6 labels, ≈10%). Every reference method is trained with the same Light-UNETR backbone under the CSE protocol. Best in bold. FHEAT-Seg entries are means over three random seeds (± SD).

| | LA (4 labels) | | | | Pancreas-CT (6 labels) | | | |
|---|---|---|---|---|---|---|---|---|
| **Method** | **Dice↑** | **Jac↑** | **95HD↓** | **ASD↓** | **Dice↑** | **Jac↑** | **95HD↓** | **ASD↓** |
| UA-MT | 78.34 | 65.11 | 15.42 | 4.69 | 52.41 | 36.80 | 36.34 | 11.78 |
| SASSNet | 79.69 | 67.02 | 12.44 | 3.86 | 51.31 | 36.10 | 22.04 | 7.50 |
| CAML | 79.74 | 66.94 | 14.58 | 4.46 | 57.34 | 40.94 | 28.41 | 8.86 |
| MLRP | 81.27 | 69.36 | 10.70 | 3.12 | 67.05 | 51.44 | 13.71 | 4.39 |
| BCP | 84.76 | 73.88 | 10.22 | 2.89 | 68.93 | 53.86 | 14.01 | 4.37 |
| Light-UNETR | 89.53 | 81.15 | 6.14 | 1.88 | 73.77 | 59.44 | 11.77 | 3.76 |
| FHEAT-Seg (ours) | **90.47±0.21** | **82.65±0.35** | **4.60±0.42** | **1.78±0.15** | **78.79±0.38** | **65.53±0.52** | **7.80±0.61** | **1.84±0.22** |

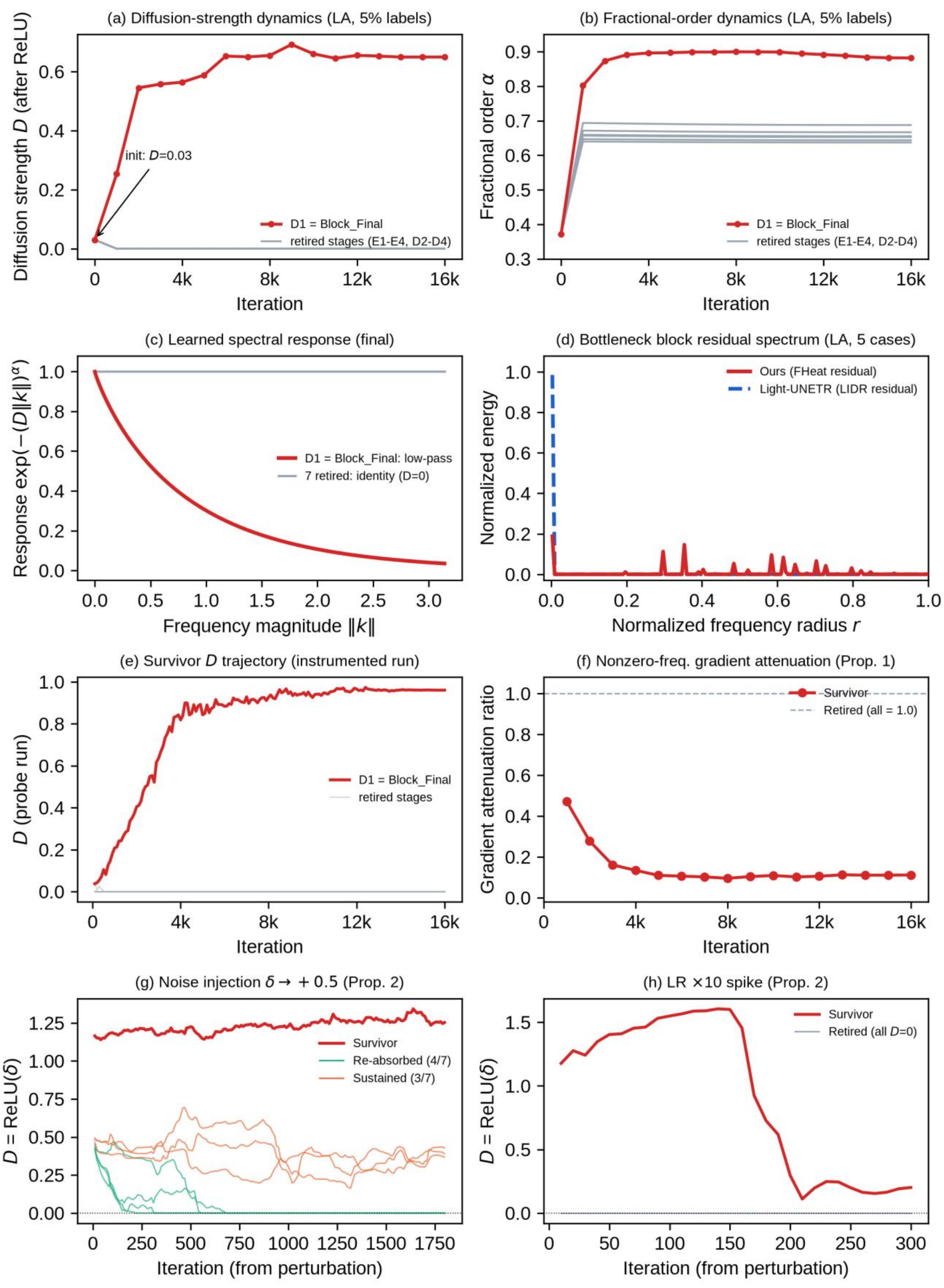

**Fig. 5.** (a, b) Training dynamics of the spectral parameters on LA at 5% labels: the sole survivor (Block_Final, red) grows monotonically while the seven retired stages (gray) collapse to $D = 0$ by the first recorded checkpoint. (c) Learned spectral response $\exp(-(D \parallel k \parallel)^{\alpha})$ at convergence: a strong low-pass at the survivor, identity elsewhere. (d) Bottleneck block residual spectrum: 33.3% of residual energy at high normalized radii ($r > 0.6$), versus 0.2% for Light-UNETR. (e) Survivor $D$ trajectory in the gradient-instrumented run. (f) Nonzero-frequency gradient attenuation ratio: the survivor deepens from $0.472$ to $0.111$; retired stages hold at $1.0000$. (g) NOISE-INJECTION PERTURBATION: 4/7 reactivated stages re-absorb to exactly $D = 0$; the remaining 3/7 sustain $D \approx 0.4$ and degrade validation Dice ($90.26 \rightarrow 87.22$). (h) Learning-rate $\times\, 10$ spike: all retired stages hold at exactly $D = 0$.

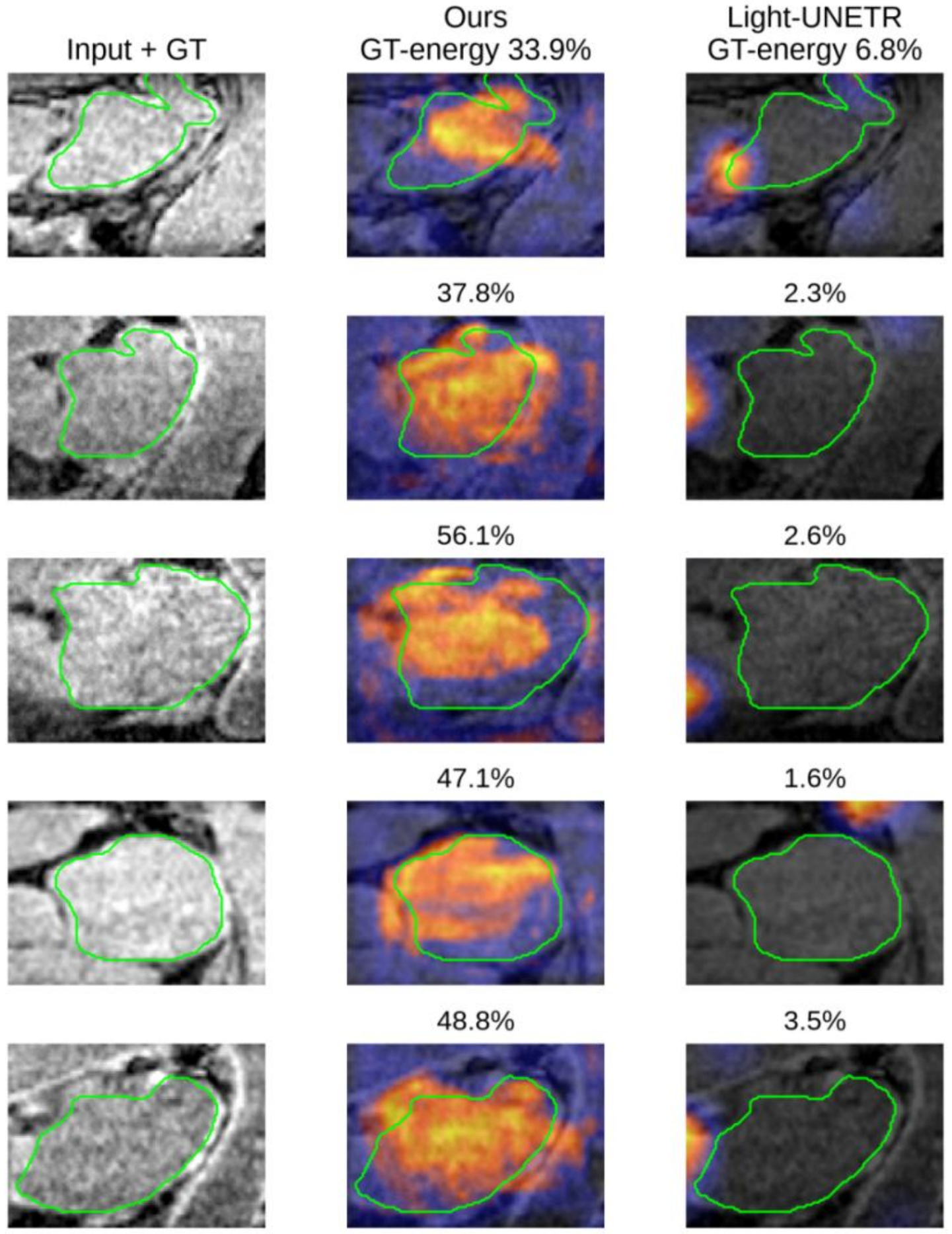

**Fig. 6.** AGR attention maps (LA, five representative cases): FHEAT-Seg concentrates 44.8% of attention energy within the target region versus 3.4% for Light-UNETR (18.4% vs. 11.6% on Pancreas-CT; 22.5% vs. 1.3% on BraTS 2019; all ratios averaged per case over the full test sets).

TABLE III. SEMI-SUPERVISED SEGMENTATION AT THE HIGHER LABEL RATE (LA: 8 LABELS, 10%; PANCREAS-CT: 12 LABELS, ≈20%; BRATS 2019: 25 LABELS, ≈10%). SAME CONVENTIONS AS TABLE II.

| | LA (8 labels) | | | | Pancreas-CT (12 labels) | | | | BraTS 2019 (25 labels) | | | |
|---|---|---|---|---|---|---|---|---|---|---|---|---|
| **Method** | **Dice↑** | **Jac↑** | **95HD↓** | **ASD↓** | **Dice↑** | **Jac↑** | **95HD↓** | **ASD↓** | **Dice↑** | **Jac↑** | **95HD↓** | **ASD↓** |
| UA-MT | 82.53 | 70.73 | 11.80 | 3.84 | 67.28 | 51.99 | 14.73 | 4.12 | 74.13 | 62.45 | 16.23 | 4.42 |
| SASSNet | 83.51 | 72.19 | 9.93 | 2.78 | 67.60 | 52.23 | 18.06 | 6.39 | 78.79 | 67.96 | 13.65 | 3.96 |
| CAML | 82.36 | 70.56 | 9.79 | 3.08 | 68.76 | 52.86 | 11.96 | 3.92 | 78.35 | 67.02 | 13.34 | 3.85 |
| MLRP | 82.92 | 71.18 | 10.24 | 2.96 | 71.80 | 57.24 | 8.84 | 2.45 | 78.10 | 67.51 | 12.50 | 3.79 |
| BCP | 86.81 | 76.94 | 8.91 | 2.49 | 73.15 | 58.56 | 8.61 | 2.41 | 77.06 | 66.02 | 18.10 | 5.33 |
| Light-UNETR | 90.50 | 82.74 | 5.82 | 1.83 | 78.50 | 65.27 | 8.35 | 2.24 | 79.73 | 68.76 | 11.65 | 2.25 |
| FHEAT-Seg (ours) | **91.85±0.16** | **84.97±0.24** | **4.30±0.35** | **1.44±0.12** | **81.07±0.29** | **68.64±0.41** | **8.69±0.48** | **1.43±0.14** | **81.90±0.33** | **71.53±0.45** | **10.27±0.72** | **2.06±0.19** |

*D. Fully Supervised Comparison and Efficiency*

With no semi-supervised component involved, the fully supervised comparison isolates the backbone itself. Table IV reports results on LA, Pancreas-CT, and BraTS 2019 against three SAM-family foundation models and seven segmentation networks. FHEAT-Seg improves over Light-UNETR on LA (92.05 vs. 91.58 Dice), Pancreas-CT (83.30 vs. 80.45), and BraTS 2019 (84.28 vs. 79.93) at 27% fewer parameters and 79% fewer FLOPs (0.975M / 0.90G vs. 1.34M / 4.29G); the parenthesized with-DCT costs show that the architecture alone already undercuts Light-UNETR's FLOPs by 72% before the learned bypass. FHEAT-Seg-L outperforms Light-UNETR-L on all three benchmarks (93.09 vs. 92.22, 85.11 vs. 84.30, 87.19 vs. 84.32) at lower cost (2.851M / 55.75G vs. 3.93M / 59.45G). Across the fourteen metric and efficiency columns of Table IV, our two variants together rank first on twelve. The larger BraTS margin is consistent with the spectral division of labor: multi-region tumor boundaries are high-frequency structures, exactly what the FHEAT blocks amplify (Section IV-B). Because this comparison involves no AGR or SMC component, the consistent margin of FHEAT-Seg-L over Light-UNETR-L indicates that the learned spectral allocation is not a byproduct of the semi-supervised protocol but a property of the backbone's optimization landscape under ordinary supervision; Section V returns to this point.

TABLE IV. FULLY SUPERVISED SEGMENTATION RESULTS (DICE/JAC IN %; 95HD/ASD IN VOXELS) ON LA, PANCREAS-CT, AND BRATS 2019, WITH EFFICIENCY ACCOUNTING. REFERENCE ROWS ARE AS REPORTED IN [2], WHERE FLOPS ARE COMPUTED AT A SPATIAL SIZE OF (112, 112, 80). BEST IN BOLD.(SECTION III-E).

| Method | LA | | | | Pancreas-CT | | | | BraTS 2019 | | | | FLOPs(G)↓ | Params(M)↓ |
|---|---|---|---|---|---|---|---|---|---|---|---|---|---|---|
| | Dice↑ | Jac↑ | 95HD↓ | ASD↓ | Dice↑ | Jac↑ | 95HD↓ | ASD↓ | Dice↑ | Jac↑ | 95HD↓ | ASD↓ | | |
| Foundation models | | | | | | | | | | | | | | |
| MedSAM2 [32] | 80.44 | 67.67 | 21.59 | 5.28 | 69.01 | 53.22 | 22.23 | 5.31 | 83.08 | 74.02 | 17.21 | 6.36 | 2,259.2 | 38.96 |
| Efficient-MedSAM2 [32] | 78.61 | 65.39 | 21.82 | 6.02 | 68.92 | 53.06 | 18.83 | 5.55 | 82.29 | 71.06 | 11.13 | 6.01 | 1,922.0 | 34.10 |
| Med-FastSAM [33] | 83.16 | 69.53 | 19.02 | 5.10 | 70.30 | 59.71 | 19.72 | 4.88 | 82.06 | 71.58 | 10.84 | 5.03 | 1,784.9 | 14.48 |
| Segmentation networks | | | | | | | | | | | | | | |
| nnU-Net [5] | 70.62 | 63.93 | 25.45 | 1.73 | 76.76 | 64.21 | 16.03 | 1.14 | 77.29 | 67.40 | 12.28 | 2.09 | 227.88 | 13.82 |
| SegResNet [34] | 84.40 | 76.45 | 10.50 | 1.93 | 81.37 | 69.21 | 7.76 | 1.21 | 77.67 | 67.17 | 12.13 | 2.02 | 229.87 | 10.64 |
| TransBTS [35] | 69.76 | 59.79 | 23.45 | 4.61 | 76.43 | 63.09 | 11.82 | 1.62 | 79.24 | 69.42 | 9.61 | 1.79 | 161.14 | 30.61 |
| Swin UNETR [7] | 87.86 | 79.54 | 7.08 | 2.00 | 78.83 | 66.13 | 9.96 | 1.20 | 77.68 | 68.11 | 12.36 | 1.75 | 591.87 | 61.99 |
| UNETR++ [36] | 91.12 | 83.89 | 5.56 | 1.52 | 83.32 | 71.99 | 5.96 | 1.05 | 79.45 | 70.00 | 11.84 | 1.69 | 81.64 | 20.33 |
| MedNeXt [37] | 92.05 | 85.35 | 4.95 | 1.36 | 83.82 | 72.50 | 4.72 | 1.16 | 82.88 | 73.75 | 9.84 | 1.76 | 65.62 | 5.57 |
| Light-UNETR [2] | 91.58 | 84.51 | 4.95 | 1.43 | 80.45 | 67.96 | 6.19 | 1.30 | 79.93 | 69.33 | 11.44 | 2.26 | 4.29 | 1.34 |
| Light-UNETR-L [2] | 92.22 | 85.61 | 4.64 | 1.36 | 84.30 | 73.12 | **4.36** | **1.04** | 84.32 | 74.98 | 9.15 | 1.65 | 59.45 | 3.93 |
| FHEAT-Seg (ours) | 92.05 | 85.35 | 4.52 | 1.49 | 83.30 | 71.83 | 6.18 | 1.10 | 84.28 | 74.62 | 9.65 | 1.73 | **0.90 (1.20)** | **0.975** |
| FHEAT-Seg-L (ours) | **93.09** | **87.11** | **3.99** | **1.16** | **85.11** | **74.32** | 4.42 | 1.05 | **87.19** | **78.40** | **8.17** | **1.22** | 55.75 (57.77) | 2.851 |

Table V extends the comparison to three additional fully supervised tasks. In paired comparisons against their same-family baselines, both variants achieve higher average Dice on all three benchmarks at lower computational cost: FHEAT-Seg outperforms Light-UNETR by +1.96, +1.94, and +1.09 points on MSD Task01, AbdomenCT-1K, and HNC Tumor, respectively, while FHEAT-Seg-L exceeds Light-UNETR-L by +1.38, +0.69, and +0.61 points.This consistent gain reflects a spectral division of labor across tumor sub-regions. For the most boundary-sensitive enhancing tumor (ET), the high-frequency FHEAT pathway yields the largest improvement: FHEAT-Seg reaches 78.54 Dice versus 67.01 for Light-UNETR (+11.53), and FHEAT-Seg-L reaches 78.31 versus 75.41 for Light-UNETR-L. For the low-contrast, diffusely bounded tumor core (TC), the compact model trails its paired baseline (83.41 vs. 86.89), as its single surviving low-pass operator over-smooths the already weak boundary signal. To quantify this boundary-contrast hypothesis, we computed image gradient magnitudes along 1-voxel boundary bands across all 484 MSD Task01 training volumes(Fig. 7). On T1gd, the TC boundary has a mean gradient of 129.2 versus 166.1 for ET (ratio 1.286, paired Wilcoxon $p = 8.1\times10^{-64}$), confirming TC boundaries are systematically weaker on the modality where the low-pass operator exerts its smoothing effect; the ordering reverses on FLAIR ($p = 2.5\times10^{-11}$). The large variant closes and reverses this TC gap (93.51 vs. 92.89) via its wider convolutional stem that preserves a stronger residual high-frequency pathway. For whole tumor segmentation, FHEAT-Seg-L matches Light-UNETR ' s 90.24 Dice and surpasses Light-UNETR-L's 89.69.

TABLE V. FULLY SUPERVISED SEGMENTATION DICE SCORES (%) ON MSD TASK01 [38], ABDOMENCT-1K [39], AND HNC TUMOR [40], WITH EFFICIENCY ACCOUNTING (FLOPS/PARAMS AT $96^3$). WT, ET, AND TC DENOTE WHOLE TUMOR, ENHANCING TUMOR, AND TUMOR CORE; GTVP AND GTVN DENOTE THE PRIMARY AND NODAL GROSS TUMOR VOLUMES. BEST IN BOLD.

| Method | MSD Task01 | | | | AbdomenCT-1K | | | | | HNC Tumor | | | FLOPs↓ | Params↓ |
|---|---|---|---|---|---|---|---|---|---|---|---|---|---|---|
| | WT | ET | TC | Avg | Liver | Kidney | Spleen | Pancreas | Avg | GTVp | GTVn | Avg | | |
| nnU-Net | 85.83 | 62.71 | 83.42 | 77.32 | 97.21 | 96.92 | 95.93 | 93.42 | 95.87 | 94.79 | 95.76 | 95.27 | 200.81 | 13.82 |
| SegResNet | 88.68 | 62.64 | 83.28 | 78.20 | 97.06 | 96.35 | 95.23 | 91.48 | 95.03 | 95.19 | 95.70 | 95.44 | 202.66 | 10.64 |
| TransBTS | 77.89 | 57.44 | 73.51 | 69.61 | 97.08 | 96.47 | 95.38 | 92.86 | 95.45 | 94.02 | 94.31 | 94.16 | 142.07 | 30.61 |
| Swin UNETR | 90.23 | 63.02 | 84.22 | 79.16 | 97.03 | 96.08 | 94.98 | 93.51 | 95.40 | 96.45 | 96.51 | 96.48 | 331.95 | 61.99 |
| UNETR++ | 90.07 | 64.87 | 84.52 | 79.82 | 95.64 | 95.56 | 93.99 | 91.68 | 94.22 | 95.39 | 95.74 | 95.56 | 57.51 | 20.33 |
| MedNeXt | 90.15 | 64.34 | 85.91 | 80.13 | 95.57 | 95.30 | 94.39 | 91.32 | 94.15 | 96.62 | **97.03** | 96.82 | 57.81 | 5.57 |
| Light-UNETR | **90.24** | 67.01 | 86.89 | 81.38 | 94.36 | 93.97 | 92.90 | 87.65 | 92.22 | 87.48 | 84.86 | 86.17 | 3.73 | 1.34 |
| Light-UNETR-L | 89.69 | 75.41 | 92.89 | 85.97 | 97.26 | 96.38 | 95.46 | 93.71 | 95.70 | 96.62 | 96.34 | 96.48 | 52.15 | 3.93 |
| FHEAT-Seg | 88.07 | **78.54** | 83.41 | 83.34 | 96.12 | 95.20 | 94.45 | 90.87 | 94.16 | 88.22 | 86.29 | 87.26 | **0.79(1.06)** | **0.975** |
| FHEAT-Seg-L | **90.24** | 78.31 | **93.51** | **87.35** | **97.82** | **97.02** | **96.22** | **94.51** | **96.39** | **97.21** | 96.96 | **97.09** | 49.15(50.93) | 2.851 |

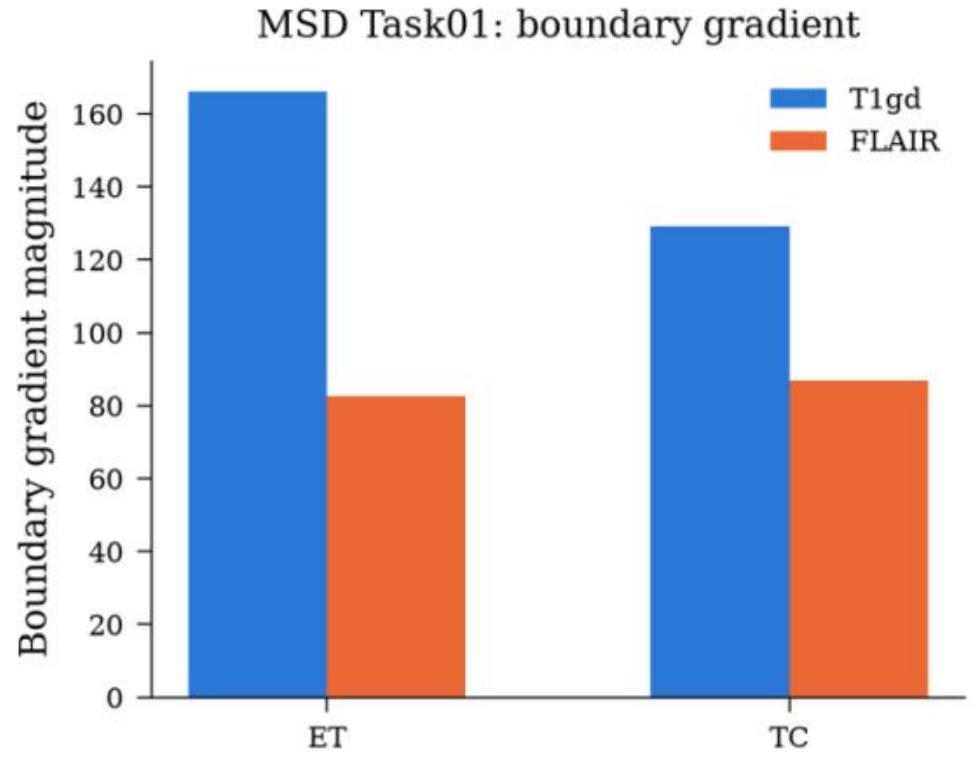

Fig. 7. MSD Task01 boundary gradient contrast: ET vs. TC on T1gd and FLAIR. The TC boundary gradient is 22.2% lower than ET on T1gd (129.2 vs. 166.1, paired Wilcoxon $p = 8.1\times10^{-64}$).

### *E. Ablation Studies*

All ablations use the LA 5% semi-supervised protocol with every hyperparameter fixed, following the single-setting ablation convention of the reference protocol [2]; each row changes exactly one factor (Table VI). As in that protocol, 10% of the unlabeled training set of LA is held out as a validation set for model selection, and performance is reported on the test set. Relative to the Light-UNETR backbone (89.53%), equipping the stages with FHEAT and an equal-capacity GELU feed-forward mixer raises Dice to 90.15%; replacing the mixer with KAN3D further raises it to 90.47% (+0.32 Dice, +0.53 Jac at nearly identical parameter counts, 0.969M vs. 0.975M). The KAN mixer is therefore responsible for a measurable, isolated share of the gain. The role of spectral mixing is best read from the controlled rows: freezing every diffusion strength at zero leaves Dice essentially unchanged (90.42% vs. 90.47%), while allowing a single learnable D only at Block_Final improves both Dice and boundary quality beyond the fully learnable model (90.60% vs. 90.47%). Concentrating the spectral budget where the network itself places it is thus not merely sufficient but slightly superior, and the seven redundant learnable D's of the full model are unnecessary. This echoes the lottery-ticket structure [41], [42], except that the sparse subnetwork here emerges from ordinary gradient descent on an identity-containing gate, rather than from post hoc pruning or a search over initializations.Expanding the KAN ratio to 2 gives 90.30%, and doubling the KAN learning-rate multiplier to 0.1 gives 89.45% with every D driven to zero. Performance degrades smoothly across this band and remains competitive even when the spectral branch is entirely inactive.

TABLE VI. ABLATIONS ON LA 5% SEMI-SUPERVISED SEGMENTATION. IDENTICAL PROTOCOL; ONE FACTOR VARIED PER ROW. THE LAST FOUR ROWS ARE CONTROLLED EXPERIMENTS, NOT CANDIDATE MODELS.

| Configuration | Dice↑ | Jac↑ | 95HD↓ | ASD↓ |
|---|---|---|---|---|
| CSE-Light-UNETR backbone | 89.53 | 81.15 | 6.14 | 1.88 |
| + FHEAT + equal-capacity FFN | 90.15 | 82.12 | 4.66 | **1.75** |
| + FHEAT + KAN3D (full model) | 90.47 | 82.65 | 4.60 | 1.78 |
| KAN ratio 2 | 90.30 | 82.37 | 4.80 | 1.83 |
| KAN lr multiplier 0.1 (all D=0) | 89.45 | 81.02 | 5.99 | 1.91 |
| All D frozen at 0 | 90.42 | 82.55 | 5.14 | 1.87 |
| D learnable at Block_Final only | **90.60** | **82.87** | **4.59** | 1.78 |

### *F. Qualitative Results*

Fig. 8 and Fig. 9 show segmentation results of three test cases per dataset under the 5% (LA) and 10% (Pancreas-CT) settings, comparing five SSL methods, the Light-UNETR backbone, and FHEAT-Seg against the ground truth; all predictions come from models re-trained under the shared backbone and protocol. FHEAT-Seg preserves thin structures (papillary muscles on LA; the pancreatic tail on Pancreas-CT) that the competing methods blur or miss, consistent with the high-frequency residual analysis above.

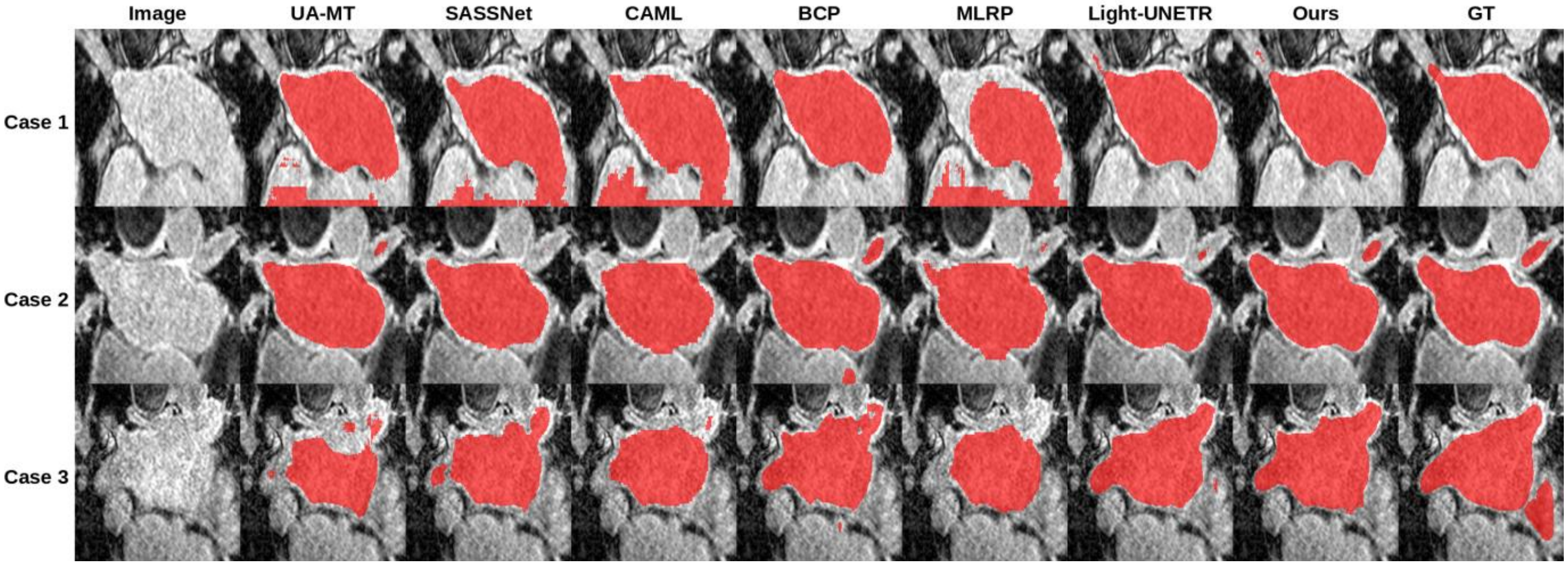

**Fig. 8.** Qualitative comparison on LA (5% labels, three test cases): Image | UA-MT | SASSNet | CAML | BCP | MLRP | Light-UNETR | FHEAT-Seg (ours) | GT.

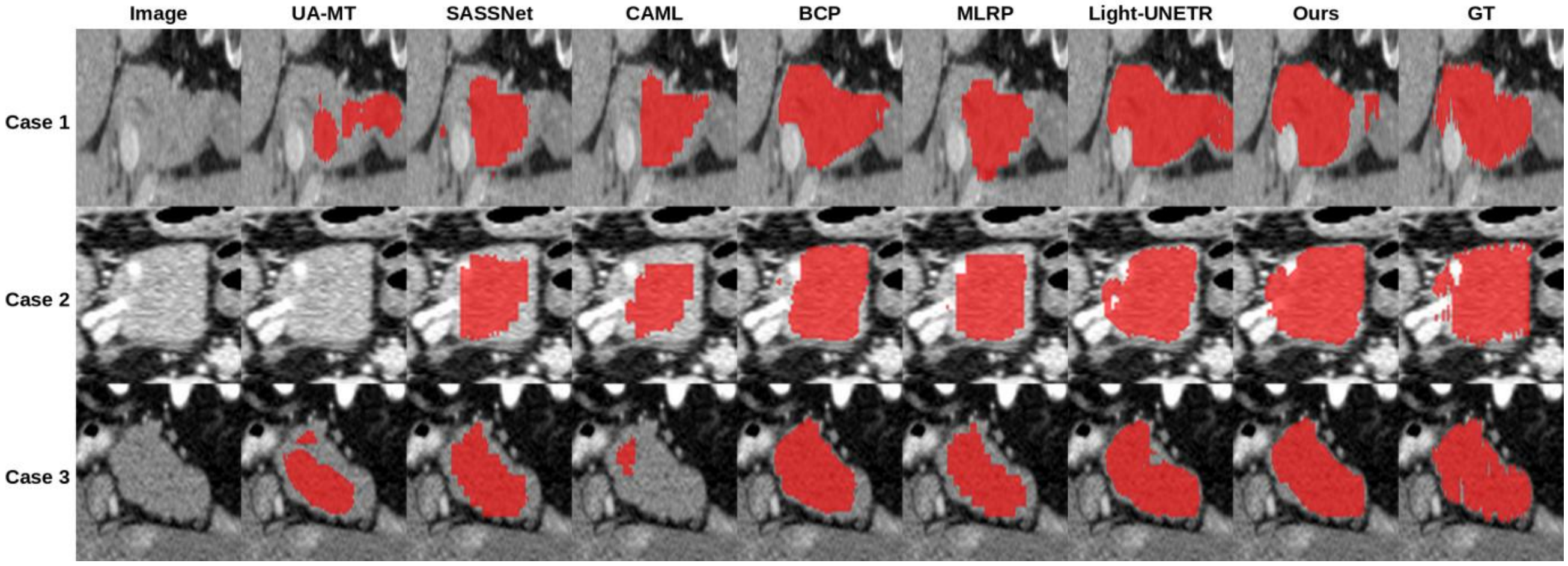

**Fig. 9.** Qualitative comparison on Pancreas-CT (10% labels, three test cases), same column layout as Fig. 8.

## V. DISCUSSION

The results invite a closer look at why a fractional spectral family suits this problem. Two properties stand out. First, the family interpolates continuously between the identity and a strong low-pass operator, so the amount of global mixing at each stage becomes a quantitative learning outcome rather than a binary architectural choice. Second, the shape of the learned response is informative: in every semi-supervised model the surviving operator saturates near $\alpha = 0.9$, the sharpest low-pass the admissible range allows, while the $\alpha$ values of the retired stages (0.60–0.75) never come into play. Formally, optimizing $\alpha$ through $\alpha = 0.3 + 0.6\sigma(\theta)$ solves a bound-constrained problem $\min_{\alpha} \mathcal{L}(\alpha)$ with $\alpha \in (0.3, 0.9)$. The cap $\alpha^* = 0.9$ is a KKT point whenever the residual descent direction is non-positive there ($\partial\mathcal{L}/\partial\alpha \le 0$ at the cap), the active upper bound absorbing the remaining descent through its multiplier. The sigmoid map makes this boundary point attainable and stable, since $d\alpha/d\theta = 0.6\sigma(\theta)(1-\sigma(\theta)) \to 0$ as the target grows, so any unconstrained preference for $\alpha > 0.9$ is absorbed at the cap. The observed saturation is therefore the signature of a boundary optimum rather than an interior one, not evidence that 0.9 is intrinsically special. Confirming that the unconstrained target lies beyond the range requires widening the parameterization: the concrete test is to replace the bounded map $\alpha = 0.3 + 0.6\sigma(\theta)$ with an unbounded one such as $\alpha = 0.3 + \mathrm{softplus}(\theta)$ and check whether the learned value moves past 0.9 or returns to it, which we leave to future work. Qualitatively, on the family $m(\rho) = \exp(-\tau\rho^{\alpha})$, increasing $\alpha$ steepens the roll-off in frequency (closer to an ideal low-pass) and lightens the spatial tails of the corresponding kernel, approaching a Gaussian as $\alpha \to 2$; saturating at the upper bound of our range means the loss wants a sharper spectral cutoff than the box allows, not that 0.9 is a physical constant.

The location of the surviving operator is equally informative. It is the final decoder stage (Block_Final), which operates at high resolution immediately before the segmentation head and whose output additionally yields the attention map used for region selection; a strong low-pass aggregation at this position denoises the features that determine both the object boundaries and the attention map. That the network requests global aggregation at exactly this stage, and nowhere else, reads as a sensible, learned division of labor rather than an artifact of the parameterization. The location is also consistent with the spectral bias of gradient descent, which fits the low-frequency components of a target mapping first [43], [44]: residual high-frequency detail is supplied locally by the convolutional path of every block, so the single operation that benefits most from an explicit learnable filter is low-pass aggregation immediately before the segmentation head and the attention read-out. The fully learnable model arrives at this block when free to choose among all eight, and the Block_Final-only control of Table VI shows that the same choice is also sufficient. In information-theoretic terms, the all-frozen control of the same table ( Dice with every D pinned at zero) indicates that the retired stage's spectral filtering carries negligible additional mutual information with the segmentation objective once the surrounding convolutional path is in place.

The relationship between backbone and protocol also deserves care. The fully supervised results of Table IV involve no AGR or SMC component and therefore isolate the backbones contribution from the semi-supervised protocol. Under semi-supervised training, the surviving operator coincides with the block that produces the attention map consumed by AGR, and its low-pass output is consistently associated with more concentrated attention maps (Section IV-B). Because the attention map is computed with a stop-gradient (Section III-D), we read this association as two products of the same learned aggregation, boundary-aware decoding and informative region selection, rather than as a causal loop from the protocol to the operator; the Block_Final-only control provides the testable check. One caveat remains: part of the semi-supervised gain may still be mediated by the improved attention quality, and the fully supervised comparison is our control for that channel.

Finally, the medical-imaging relevance of the spectral design should be stated explicitly. Volumetric anatomy is rich in diagnostically critical high-frequency structure (thin walls, papillary muscles, vessel-like tails, and subtle tissue interfaces) that volumetric Dice under-weights but that determines clinical usefulness. The residual-spectrum analysis (Section IV-B) shows that the FHEAT blocks inject two order of magnitude more high-frequency detail than the low-pass-only attention of the reference backbone, and the qualitative results confirm that this translates into preserved thin structures and improved boundary distances (95HD/ASD), the metrics most sensitive to exactly those features. The fractional heat kernel is therefore not a generic mixer transplanted from natural-image vision: its learnable, frequency-selective response lets a lightweight backbone devote its limited capacity to the high-frequency anatomical detail that medical segmentation demands, while retiring spectral computation that the data do not need.

Limitations. Three points bound the scope of the present evidence. (i) The fractional order $\alpha$ is optimized over a bounded range $(0.3, 0.9)$; the observed saturation at the upper bound is a KKT point of that bound, not evidence that $0.9$ is a physically preferred order, and we have not yet run the unbounded softplus parameterization that would confirm whether the unconstrained optimum lies beyond it. (ii) The absorbing-zero mechanism of Proposition 2 is specific to the $D = \mathrm{ReLU}(\delta)$ parameterization under decoupled weight decay; it is a deliberate design choice that makes retirement exact and compilable, not a claim that every reasonable reparameterization of the diffusion strength would sparsify the same way. (iii) The argument for site-selection is closed without a checkpoint measurement: Proposition 1 derives the fractional Dirichlet pairing and its sign test, Fig. 3(b) confirms on a controlled synthetic ground truth that the sign of the pairing at initialization predicts the surviving site, Table VI is a ranking consistent with that test (Block_Final useful) rather than a numerical evaluation of the pairing, and Proposition 2 together with Fig. 5(e)–(h) explains why a stage with an unfavorable sign stays at exactly zero once reached. What we have not yet reported is the pairing's measured value at Block_Final on a saved medical checkpoint — a single forward/backward pass at $D = 0$ per site, no retraining — which is a further, direct measurement of the theory, not a piece the present argument is missing. Relatedly, Theorem 1's semigroup collapse is exact only for gates that share both resolution and fractional order; it certifies the redundancy already built into each stage by parameter sharing (Section III-A) and should not be read as claiming that the eight stage-level gates of FHEAT-Seg are themselves interchangeable, since they differ in both respects.

## VI. Conclusion

This paper asks where global spectral mixing should be spent in a lightweight 3D segmentation backbone, and shows that the answer is not a design choice but the output of an exact operator theory. FHCO is a two-parameter Neumann semigroup: same-resolution, same-order instances compose exactly under addition of a natural diffusion time τ = D^α (Theorem 1), so any stack of such gates is equivalent to a single learned-strength Sobolev regularizer, or to no regularizer at all when τ = 0. This redundancy alone does not explain sparsity, since weight decay applied to truly interchangeable gates favors spreading the diffusion budget rather than concentrating it (Remark 3). We therefore derive the actual selection rule: the first variation of the loss in τ is a fractional Dirichlet pairing between a stage's incoming feature and incoming gradient, and a stage is retained, to first order, if and only if that pairing is positive (Proposition 1); ReLU parameterization and weight decay then compile an unfavorable sign into an exact, deployable zero (Proposition 2). Table VI agrees with the ranking this theory predicts, and a fully controlled, non-medical simulation (Fig. 3) confirms both the non-sparsifying nature of pure redundancy and the sign-predicts-survival mechanism on a synthetic ground truth where the answer is known by construction.

Instantiated as FHEAT-Seg on a lightweight U-shaped backbone, this theory has concrete empirical consequences. Under scarce labels, training drives the diffusion strength to zero in all but one of eight stages, and the survivor is consistently the decoder block that also produces the semi-supervised attention map; the resulting inference graph, at 0.975M parameters and 0.90G FLOPs (79% fewer FLOPs than Light-UNETR), achieves the best semi-supervised segmentation among the compared methods under the shared CSE protocol on LA, Pancreas-CT, and BraTS 2019, and outperforms Light-UNETR-L under full supervision at lower cost. Two further results support the framework's generality as an operator-level claim rather than a Light-UNETR-specific one. First, the spectral sparsification phenomenon (unique surviving stage, exact-zero retirement) reproduces on a structurally different backbone: a 3D SegFormer with self-attention replaced by FHEAT gates, trained under the identical CSE protocol on LA at 5% labels, achieves Dice = 0.9002 vs. 0.8817 for the plain control, with 7/8 stages converging to exactly D = 0. Second, the learned allocation's optimality is confirmed by perturbation: reactivating retired stages degrades validation Dice, while the absorbing zero is stable under learning-rate spikes. The broader message is that the computational budget of a lightweight network, how much global mixing and where, is not a hyperparameter to be tuned but a quantity that an exact operator theory predicts and gradient descent discovers.

Code is released at an anonymized repository for review and will be made publicly available upon acceptance.

## References

[1] Z. Xiong et al., "A global benchmark of algorithms for segmenting the left atrium from late gadolinium-enhanced cardiac magnetic resonance imaging," Med. Image Anal., vol. 67, Oct. 2021, Art. no. 101832.

[2] X. Liu, Z. Chen, W. Li, C. Li, and Y. Yuan, "Harnessing lightweight transformer with contextual synergic enhancement for efficient 3D medical image segmentation," IEEE Trans. Pattern Anal. Mach. Intell., vol. 48, no. 3, pp. 3852–3867, Mar. 2026.

[3] O. Ronneberger, P. Fischer, and T. Brox, "U-Net: Convolutional networks for biomedical image segmentation," in Proc. MICCAI, 2015, pp. 234–241.

[4] F. Milletari, N. Navab, and S.-A. Ahmadi, "V-Net: Fully convolutional neural networks for volumetric medical image segmentation," in Proc. 3DV, 2016, pp. 565–571.

[5] F. Isensee et al., "nnU-Net: A self-configuring method for deep learning-based biomedical image segmentation," Nat. Methods, vol. 18, no. 2, pp. 203–211, Feb. 2021.

[6] J. M. J. Valanarasu, P. Oza, I. Hacihaliloglu, and V. M. Patel, "Medical transformer: Gated axial-attention for medical image segmentation," in Proc. MICCAI, 2021, pp. 36–46.

[7] A. Hatamizadeh, V. Nath, Y. Tang, D. Yang, H. Roth, and D. Xu, "Swin UNETR: Swin transformers for semantic

segmentation of brain tumors in MRI images," in Proc. MICCAI BrainLes Workshop, 2022, pp. 272–284.
[8] L. Yu, S. Wang, X. Li, C.-W. Fu, and P.-A. Heng, "Uncertainty-aware self-ensembling model for semi-supervised 3D left atrium segmentation," in Proc. MICCAI, 2019, pp. 605–613.
[9] Y. Bai, D. Chen, Q. Li, W. Shen, and Y. Wang, "Bidirectional copy-paste for semi-supervised medical image segmentation," in Proc. CVPR, 2023, pp. 11514–11524.
[10] Y. Wu, Z. Wu, Q. Wu, Z. Ge, and J. Cai, "Exploring smoothness and class-separation for semi-supervised medical image segmentation," in Proc. MICCAI, 2022, pp. 34–43.
[11] S. Gao, Z. Zhang, J. Ma, Z. Li, and S. Zhang, "Correlation-aware mutual learning for semi-supervised medical image segmentation," in Proc. MICCAI, 2023, pp. 98–108.
[12] J. Su, Z. Luo, S. Lian, D. Lin, and S. Li, "Mutual learning with reliable pseudo label for semi-supervised medical image segmentation," Med. Image Anal., vol. 94, Apr. 2024, Art. no. 103111.
[13] J. H. Lee-Thorp, J. Ainslie, I. Eckstein, and S. Ontañón, "FNet: Mixing tokens with Fourier transforms," in Proc. NAACL-HLT, 2022, pp. 4296–4313.
[14] Y. Rao, W. Zhao, Z. Zhu, J. Lu, and J. Zhou, "Global filter networks for image classification," in Proc. NeurIPS, 2021.
[15] J. Guibas, M. Mardani, Z. Li, A. Tao, A. Anandkumar, and B. Catanzaro, "Adaptive Fourier neural operators: Efficient token mixers for transformers," in Proc. ICLR, 2022.
[16] Z. Liu et al., "KAN: Kolmogorov–Arnold networks," arXiv:2404.19756, 2024.
[17] X. Yang and X. Wang, "Kolmogorov–Arnold transformer," in Proc. ICLR, 2025.
[18] H. Yang et al., "TK-Mamba: Marrying KAN with Mamba for text-driven 3D medical image segmentation," arXiv:2505.18525, 2025.
[19] Y. Hong, L. Zhang, X. Ye, and J. Mo, "SlimFormer-3D: A layer-adaptive lightweight transformer for efficient 3D medical image segmentation," in Proc. MICCAI, 2025, pp. 537–546.
[20] A. Dosi, S. Mondal, R. C. Ghosh, M. Brescia, and G. Longo, "Less is more: AMBER-AFNO—A new benchmark for lightweight 3D medical image segmentation," Expert Syst. Appl., vol. 324, 2026, Art. no. 132518.
[21] H. Liu, K. Simonyan, and Y. Yang, "DARTS: Differentiable architecture search," in Proc. ICLR, 2019.
[22] C. Louizos, M. Welling, and D. P. Kingma, "Learning sparse neural networks through $L_0$ regularization," in Proc. ICLR, 2018.
[23] G. Huang, Y. Sun, Z. Liu, D. Sedra, and K. Q. Weinberger, "Deep networks with stochastic depth," in Proc. ECCV, 2016, pp. 646–661.
[24] N. Shazeer et al., "Outrageously large neural networks: The sparsely-gated mixture-of-experts layer," in Proc. ICLR, 2017.
[25] J. Bai and X.-C. Feng, "Fractional-order anisotropic diffusion for image denoising," IEEE Trans. Image Process., vol. 16, no. 10, pp. 2492–2502, Oct. 2007.
[26] D. Ha, A. Dai, and Q. V. Le, "HyperNetworks," in Proc. ICLR, 2017.
[27] P. R. Stinga and J. L. Torrea, "Extension problem and Harnack's inequality for some fractional operators," Comm. Partial Differential Equations, vol. 35, no. 11, pp. 2092–2122, 2010.
[28] W. L. Hare and A. S. Lewis, "Identifying active constraints via partial smoothness and prox-regularity," J. Convex Anal., vol. 11, no. 2, pp. 251–266, 2004.
[29] H. R. Roth et al., "Data from pancreas-CT," The Cancer Imaging Archive (TCIA), 2016.
[30] J. Xiang, P. Qiu, and Y. Yang, "FUSSNet: Fusing two sources of uncertainty for semi-supervised medical image segmentation," in Proc. MICCAI, 2022, pp. 481–491.
[31] B. H. Menze et al., "The multimodal brain tumor image segmentation benchmark (BRATS)," IEEE Trans. Med. Imag., vol. 34, no. 10, pp. 1993–2024, Oct. 2015.
[32] J. Ma et al., "MedSAM2: Segment anything in 3D medical images and videos," arXiv:2504.03600, 2025.
[33] Y. Luo, Q. Xu, J. Feng, G. Qian, and W. Duan, "Med-FastSAM: Improving transfer efficiency of SAM to domain-generalised medical image segmentation," in Proc. NeurIPS Workshop Advancements Med. Found. Models, 2024.
[34] A. Myronenko, "3D MRI brain tumor segmentation using autoencoder regularization," in Proc. MICCAI BrainLes Workshop, 2018, pp. 311–320.
[35] C. Wang, C. Chen, M. Ding, H. Yu, S. Zha, and J. Li, "TransBTS: Multimodal brain tumor segmentation using transformer," in Proc. MICCAI, 2021, pp. 109–119.
[36] A. Shaker, M. Maaz, H. Rasheed, S. Khan, M.-H. Yang, and F. S. Khan, "UNETR++: Delving into efficient and accurate 3D medical image segmentation," IEEE Trans. Med. Imag., vol. 43, no. 9, pp. 3377–3390, Sep. 2024.
[37] S. Roy, J. M. Wolterink, and I. Išgum, "MedNeXt: Transformer-driven scaling of ConvNets for medical image segmentation," in Proc. MICCAI, 2023, pp. 405–415.
[38] M. Antonelli et al., "The medical segmentation decathlon," Nat. Commun., vol. 13, no. 1, Jul. 2022, Art. no. 4128.
[39] J. Ma et al., "AbdomenCT-1K: Is abdominal organ segmentation a solved problem?" IEEE Trans. Pattern Anal. Mach. Intell., vol. 44, no. 10, pp. 6695–6714, Oct. 2022.
[40] K. A. Wahid, C. Dede, M. A. Naser, and C. D. Fuller, Head and Neck Tumor Segmentation for MR-Guided Applications: First MICCAI Challenge. Cham, Switzerland: Springer, 2025.
[41] S. Han, J. Pool, J. Tran, and W. Dally, "Learning both weights and connections for efficient neural networks," in Proc. NeurIPS, 2015, pp. 1135–1143.
[42] J. Frankle and M. Carbin, "The lottery ticket hypothesis: Finding sparse, trainable neural networks," in Proc. ICLR, 2019.
[43] N. Rahaman et al., "On the spectral bias of neural networks," in Proc. ICML, 2019, pp. 5301–5310.
[44] Z.-Q. J. Xu, Y. Zhang, T. Luo, Y. Xiao, and Z. Ma, "Frequency principle: Fourier analysis sheds light on deep neural networks," CSIAM Trans. Appl. Math., vol. 1, no. 3, pp. 330–352, 2020.